\documentclass[sigconf]{acmart}
\usepackage{mdframed}
\usepackage{paralist}
\usepackage{hyperref}
\usepackage{amsmath}
\usepackage{booktabs}
\usepackage{tabularx}
\usepackage{caption}
\usepackage{subcaption}
\usepackage{censor}
\usepackage{xspace}

\def\VexCategory{\textsc{VEXCategory}\xspace}
\def\VexJustification{\textsc{VEXJustification}\xspace}
\def\Internalcomment{\textsc{InternalComment}\xspace}
\def\Customercomment{\textsc{CustomerComment}\xspace}

\AtBeginDocument{%
  }

\begin{document}

\title{CVE-LLM : Automatic vulnerability evaluation in medical device industry using large language models}




\author{Rikhiya Ghosh}
\authornote{Digital Technology and Innovation, Siemens Healthineers USA}
\email{rikhiya.ghosh@siemens-healthineers.com}

\author{OlaDimeji Farri}
\authornotemark[1]

\author{Hans-Martin von Stockhausen}
\authornote{Cybersecurity, Siemens Healthineers Germany}

\author{George Marica Vasile}
\authornote{Corporate Technology, Siemens AG Romania}
\author{Martin Schmitt}
\authornotemark[2]

\renewcommand{\shortauthors}{Ghosh et al.}

\begin{abstract}
 The healthcare industry is currently experiencing an unprecedented wave of cybersecurity attacks, impacting millions of individuals \cite{idtheftreport}. With the discovery of thousands of vulnerabilities each month, there is a pressing need to drive the automation of vulnerability assessment processes for medical devices, facilitating rapid mitigation efforts. Generative AI systems have revolutionized various industries, offering unparalleled opportunities for automation and increased efficiency\cite{naveed2023comprehensive}. This paper presents a solution leveraging Large Language Models (LLMs) to learn from historical evaluations of vulnerabilities for the automatic assessment of vulnerabilities in the medical devices industry. This approach is applied within the portfolio of a single manufacturer, taking into account device characteristics, including existing security posture and controls. The primary contributions of this paper are threefold. Firstly, it provides a detailed examination of the best practices for training a vulnerability Language Model (LM) in an industrial context. Secondly, it presents a comprehensive comparison and insightful analysis of the effectiveness of Language Models in vulnerability assessment. Finally, it proposes a new human-in-the-loop framework to expedite vulnerability evaluation processes.
\end{abstract}




\maketitle

\section{Introduction}
 
A software vulnerability is a weakness or flaw in a software system that can be exploited by attackers to compromise the integrity, availability, or confidentiality of the system or the data it processes. These vulnerabilities can exist at various levels of a software system -- code, design, configuration, or operation of the software. Examples of software vulnerabilities include buffer overflows, SQL injection, cross-site scripting (XSS), and insecure deserialization. There are several regulatory authorities, non-profits and companies that regularly update the growing list of software vulnerabilities. The most prominent of these organizations include MITRE corporation \footnote{\url{https://www.mitre.org/}} which, in conjunction with National Institute of Standards and Technology (NIST) and National Vulnerability Database (NVD) \footnote{\url{https://nvd.nist.gov/}}, maintain the Common Vulnerabilities and Exposures (CVE)\cite{mitre2005common} database. The CVE database is a system for uniquely identifying and cataloging known vulnerabilities in software and hardware. Each CVE entry includes a standardized identifier (e.g., CVE-2022-12345), a brief description of the vulnerability, and any relevant references or links to additional information. The goal of CVE is to provide a common language for discussing vulnerabilities and to facilitate the sharing of information about them among security professionals, vendors, and the general public. There are around 250K vulnerabilities at present, and it grows by thousands every month.

Medical devices are prone to various security threats that can compromise patient safety, privacy, the integrity of medical data, and the availability of diagnostic or treatment devices. Some common vulnerability threats include 
\begin{inparaenum}[(a)]
    \item software vulnerabilities, like buffer overflow, injection attacks, authentication bypass etc,
    \item issues with proper encryption,
    \item weak authentication measures,
    \item third party component-related vulnerabilities,
    \item use of unsupported legacy systems that are vulnerable but unpatchable,
    \item insecurely implemented/operated wireless connectivity such as missing authentication and weak encryption, and
    \item lack of adequate security controls.
\end{inparaenum}
Recent studies\cite{idtheftreport} show that healthcare industry is the most vulnerable sector with 809 data compromises and 56 million victims in 2023, which is 136\% increase from 2022. Additionally, healthcare data breach cost is extremely high, estimated to be 4.35 million USD by 2022. With the fast growing number of vulnerabilities and the threats that the healthcare industry faces, there needs to be faster and automated detection and mitigation of vulnerabilities. Medical Device Manufacturers (MDMs) are required by FDA to analyze all vulnerabilities. Most vulnerability assessment procedures being manual, this leads to delays in processing the vulnerabilities that pose actual threats to the systems.

One way of describing the severity of a CVE is using the Common Vulnerability Scoring System (CVSS)\cite{mell2006common}, developed and maintained by Forum of Incident Response and Security Teams (FIRST)\footnote{\url{https://www.first.org/}}. CVSS describes the severity of a CVE as a single vector that is created by several metrices describing the exploitability of the vulnerability, the impact on a given system, the current confirmation level and available fixes, and any aspects of the environment that may be available to contain the CVE. The CVSS metrics are divided into Base, Temporal and Environmental metrics. Base metrics reflect the intrinsic characteristics of a vulnerability and do not change over time or based on the environment in which the vulnerability exists. The temporal metric describes aspects of the vulnerability that may change over time, e.g. whether it is confirmed by the vendor, a fix is available, or whether it is being actively exploited in the wild. The environmental metrics is used to adapt the severity of a CVE based on the local CIA requirements (Confidentiality, Integrity, and Availability) or existing controls of a product/system in which the CVE is deployed. This might either be a 3rd party component inside of a product or a product inside of an operator’s environment. Vulnerability management involves the three-step approach of detection, evaluation, and mitigation. In the context of this paper we discuss the evaluation of vulnerabilities in 3rd party components embedded in medical devices. Detection is performed by the vendors of those components which afterwards publish a security notification about it. Medical device manufacturers must monitor for those notifications and perform the evaluation of the component related CVE in the context of each product the component is used in. Mitigation can either be performed by the component vendor by providing a patch that is subsequently applied to the medical device or by the medical device manufacturer by adding controls to the product that contain the CVE. In fortunate scenarios, the CVE of the component does not pose a risk to the medical device and hence no mitigation is necessary but only information from the MDM towards their customers. 
The approaches to automatic vulnerability assessment involves use of machine learning methods\cite{prasad2018integrated, iorga2020early, levshun2023comparative} and more recently, Large Language Models (LLMs)\cite{yin2024multitask, lu2024grace, steenhoek2024comprehensive}.

Large Language Models are a class of machine learning models that are capable of understanding and generating human language. These models are typically based on transformer architectures and are trained on large amounts of text data to learn the statistical patterns and structures of natural language. Since the release of ChatGPT\cite{chatgpt}, use of LLMs has been ubiquitous and has been adapted for Healthcare\cite{sallam2023chatgpt}, IT\cite{wu2023autogen}, Cybersecurity\cite{zhang2024llms}, Finance\cite{li2023large}, Legal applications\cite{yang2024large}, content creation\cite{moore2023empowering} and every conceivable text-based application. The most commonly used paradigms of LLM integration in applications have been prompt engineering and retrieval-augmented generation (RAG)\cite{lewis2020retrieval}. Domain-specific applications also employ methods like domain adapted pretraining (DAPT)\cite{ge2024openagi}, supervised finetuning (SFT)\cite{vonwerra2022trl}, ensemble methods\cite{lu2023routing},  knowledge integration\cite{feng2023knowledge}, transfer learning\cite{chronopoulou2019embarrassingly}, data augmentation\cite{whitehouse2023llm}, reinforcement learning\cite{lee2023rlaif} to achieve higher performance with their respective datasets. Some of the most common applications of LLMs in cybersecurity are vulnerability detection \cite{steenhoek2024comprehensive}, automated program repair\cite{zhou2024large}, secure code generation\cite{fu2024constrained}, binary code-based similarity and recovery\cite{tan2024llm4decompile}, cybersecurity assistants \cite{zhang2024llms}, and infrastructure maintenance\cite{naito2023llm}. LLMs and vulnerability management are intricately linked, with LLMs playing vital roles in detecting, repairing, and mitigating vulnerabilities, and vulnerability management protocols help in assessing and mitigating security risks in LLM-powered applications\cite{yao2024survey}. The evolution of advanced LLM-based reasoning techniques has spurred significant research into using LLMs to rectify vulnerabilities in code and provide explanations for vulnerability detection and repair processes.

Use of LLMs for vulnerability assessment mostly focuses on detection of vulnerabilities using source code\cite{sun2024llm4vuln}. Recent work on vulnerability assessment using source code has shown the limitations that LLMs face in assessing vulnerability severity based on source code only, and report increase in performance with descriptive text and context information. Cybersecurity researchers have envisaged using a LLM which is finetuned with asset information and vulnerability-related data held by an organization to output an effective penetration test attack path \cite{llm_it}. However, there is a dearth of real-world experiments demonstrating the effectiveness of employing LLMs for vulnerability management. This has prompted us to develop a large language model that is trained using historical evaluation data containing the description of an asset and a vulnerability to perform detailed evaluation of the impact of the vulnerability on the asset. This would aid in fast detection and assessment of applicable vulnerabilities which in turn would help in faster mitigation efforts. 

In this paper, we have developed a LLM that would generate vulnerability assessments in the form of structured vectors and categories as well as detailed description of how the vulnerability impacts the system for different stakeholders. To the best of our knowledge, this is the first paper that explores using LLMs for vulnerability assessment based solely on asset and vulnerability descriptions in medical technology industry setting. We have also benchmarked our model with respect to other finetuned open source LLMs and offer insights into the best practices to train a vulnerability LLM. Finally, we have proposed a human-in-the-loop vulnerability assessment system that drastically reduces the time taken to assess each vulnerability, which in turn could benefit cybersecurity for medical devices.

\section{Problem Statement}
Starting with the US Food and Drug Adminstration's (FDA) Guidance on Postmarket management of Cybersecurity in medical devices \cite{fdareport}. MDMs are required to monitor all 3rd party software components used in medical devices for emerging vulnerabilities and evaluate them towards whether they have a negative impact on the safety and security for continued clinical use of that device. Based on the size of the product portfolio and the number of 3rd party components utilized, the amount of device specific evaluations to perform can be overwhelming for an MDM if they are performed manually and given the development of the amount of reported vulnerabilities over the last years as described in the previous chapter. Automation of evaluations is seen as an potent measure to cope with the amount of evaluations efficiently and achieving a good quality level. Furthermore, those evaluations need to be performed over the complete post-market lifetime of a device which might be up to 20 years and during which the knowledgeable experts that can perform evaluations quickly may no longer be available in the same position for a given product or even in the same company.

Vulnerability evaluation in our purview is determination of the following assessments, given the details of an asset and a notification.
\begin{enumerate}
    \item \VexCategory : Categorization of the vulnerability to signify whether the asset is 
    \begin{inparaenum}
        \item affected,
        \item not affected
    \end{inparaenum}
    by the notification.
    \item \VexJustification : This explains the \VexCategory further in case the asset is not affected by the notification. 
    \item Vector: This refers to the CVSS environmental metrics.
    \item Internal Comment: This contains a description of the problem and a solution advised for internal use.
    \item Customer Comment: This is a brief comment made about how the notification affects the asset and the solution, and is directed towards the customer.
\end{enumerate}
An asset detail typically contains organization-wide details of the asset, including the list of 3rd party components used by the asset. Notification details comprises of details of one or more vulnerabilities that constitute the notification and the components affected by it. 
The largest efficiency gain is in automatically identifying CVEs that do NOT apply to the product together with VEX based reasoning. Newly emerged CVEs that apply to a product need a first time human investigation due to the absence of design information in today's LLM.

\section{Relevant Work}

Language models (LMs) have been a pivotal area of research in natural language processing (NLP), driving significant advancements in various NLP tasks. The large data revolution in NLP was introduced by Transformer architecture \cite{devlin2018bert} by enabling parallel computation and efficient modeling of long-range dependencies in text. The architecture is constituted by an encoder-decoder design, featuring multiple layers of both self-attention and feedforward neural networks. In recent years, advancements in high-performance computing, scalability, architecture, and data availability have facilitated the evolution from models like BERT \cite{devlin2018bert} and GPT \cite{radford2018improving} to more sophisticated iterations such as GPT-4 \cite{achiam2023gpt}, Llama3 \cite{touvron2023llama}, Mixtral \cite{jiang2024mixtral}, and other powerful Large Language Models (LLMs). These models exhibit remarkable performance not only in text-based systems but also in multimodal domains. Language Models have been used extensively in vulnerability management 
\begin{inparaenum}[(a)]
    \item to determine CVSS metrics from CVE description,
    \item to establish a mapping between vulnerabilities in CVE database with the corresponding attack tactics and techniques in ATT\&CK database \cite{attack_database},
    \item for vulnerability detection, and
    \item for vulnerability repair.
\end{inparaenum} 

CVE description to CVSS metrics mapping includes both mapping to CVSS vectors and CVSS score. Mapping to CVSS score follows a linear regression using Bag-of-Words model \cite{elbaz2020fighting} or neural network model using Doc2Vec \cite{vasireddy2023cvss} method of extracting features from a document. CVSS-BERT \cite{shahid2021cvss}, on the other hand, trained different BERT models for classification of CVE descriptions to the values of different CVE vectors. Both encoder and decoder based models have been used for Vulnerability to ATT\&CK tactics and techniques. CVET \cite{ampel2021linking}, a RoBERTa\cite{liu2019roberta}-based model, classified CVE descriptions to one of the ten tactics in ATT\&CK, whereas SMET \cite{abdeen2023smet} used BERT-based textual similarity to map CVE entries to ATT\&CK techniques. Though ChatGPT-based approaches did not yield state-of-the-art results \cite{liu2023not}, VTT-LLM \cite{zhang2024vtt}, trained using various versions of the decoder Bloom model\cite{le2023bloom} with Chain-of-thought \cite{wei2022chain} instructions, incorporated relations between core concepts in CWE (Common Weakness Enumeration) \cite{christey2013common} and CAPEC (common attack patterns) databases \cite{capec_database} for CVE to ATT\&CK mapping surpassed encoder based models. Related work  \cite{yosifova2021predicting} also involves CVE Vulnerability type classification using TF-IDF and standard machine learning classifier. 

In the domain of vulnerability detection, encoder-based LLMs \cite{ameri2021cybert, yin2020apply} have been instrumental, particularly through innovative approaches such as the pretrain-and-finetune paradigm. These approaches have also leveraged specialized methods including novel pretraining strategies, the integration of other deep learning paradigms like Graph Neural Networks\cite{sewak2023crush} or Long Short-Term Memory (LSTM)\cite{hassanin2024pllm} networks, prompt tuning, program analysis, specialized calculus-based causal reasoning, and knowledge graph-based reasoning. The focus has increasingly shifted towards decoder-only LLMs\cite{zhou2024large}, where prompt tuning, finetuning, and retrieval-augmented generation systems are heavily employed for vulnerability detection.

Conversely, vulnerability repair systems primarily utilize decoder LLMs\cite{pearce2023examining}, often relying on finetuning, transfer learning, retrieval augmented generation, or reinforcement learning-based systems \cite{islam2024llm}. These systems typically operate on diverse datasets and have witnessed the emergence of innovative transformer architectures and training techniques. Furthermore, recent advancements in the field have also explored the combination of these techniques with ensemble learning methods, meta-learning approaches, and the incorporation of domain-specific knowledge for enhanced vulnerability management in complex software systems.

\section{Methodology}

We define the vulnerability assessment problem as follows:
given an asset A and notification N, an evaluation is generated.

\begin{equation*}
    \left(A,N\right)\rightarrow \mathsf{evaluation}
\end{equation*}

An evaluation can be a \VexCategory, CVSS environmental metrics, Internal comment, or Customer Comment. \VexCategory is generated along with \VexJustification, which further elucidates the categorisation of the vulnerability.
    \begin{align*}
    \mathsf{\VexCategory} \in &\left\{ \mathsf{Affected, NotAffected} \right\} \\
    \mathsf{\VexJustification} \in &\left\{ \mathsf{VulnerableComponentNotPresent},\right.\\
    &\left. \mathsf{ComponentNotPresent}, \right.\\
    &\left. \mathsf{VulnerableCodeNotInExecutePath},\right. \\
    &\left. \mathsf{VulnerableCodeCannotBeControlledByAdversary}, \right. \\
    &\left. \mathsf{InlineMitigationsAlreadyExist}, \right. \\
    &\left. \mathsf{Other, N/A} \right \}     
    \end{align*}
\Internalcomment and \Customercomment are text explanations of the vulnerability assessments.

Our methodology follows the standard domain adaptation procedure for large language models called domain adaptive pretraining (DAPT). We have used Mosaic ML's MPT-7B LLM\cite{MosaicML2023Introducing} for domain adaptation with a combination of publicly available vulnerability-related documents and organization-wide documents. We used human expert-derived evaluations for the standard vulnerabilites for the organizational assets to form an instruction tuning dataset that is used to train CVE-LLM.

\begin{figure}[h]
  \centering
  \begin{frame}{\includegraphics[width=\linewidth]{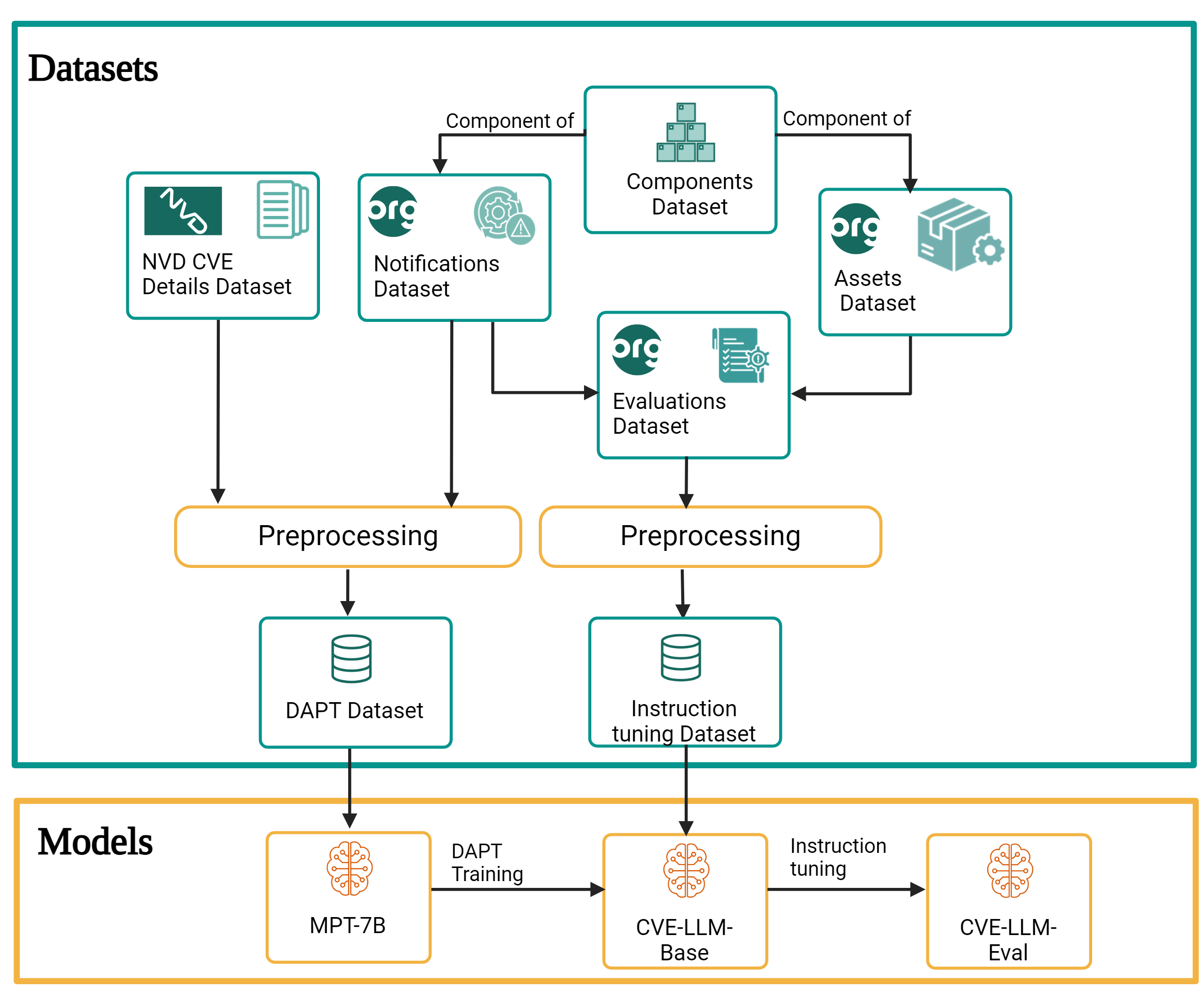}}
      
  \end{frame}
  \caption{Schematic of the Training paradigm}
  \Description{Training Schema}
\end{figure}

\subsection{Datasets}
\subsubsection{DAPT Dataset}
DAPT Dataset consists of 320K vulnerability description documents that are a combination of publicly available documents (218K) from NVD, and organization-wide vulnerability documents (102K). The organization-wide vulnerability documents describe, to various details, combination of one or more vulnerabilities and their effects and possible mitigation for the affected product.

\subsubsection{Instruction tuning Dataset}
Instruction tuning dataset is formed from three organization-wide datasets: 
\begin{inparaenum}[(a)]
    \item Assets,
    \item Notifications, and
    \item Evaluations.
\end{inparaenum} 
Assets represent all the different products released by the organization, and includes various software versions of the products. The dataset consists of asset id, product name and version, software name and version, sub-organization that the product is associated with, and list of software components that constitute the asset. 

Notifications dataset encompasses the set of all vulnerabilities that might affect one or more assets and is generally the combination of one or more CVEs. The Notifications dataset contains notification id, a description of the notification, the software components that the notification affects, the base and temporal vectors and scores for the notification, and the constituent vulnerabilities (CVEs and CWEs) for the notification. The description of a notification consists of descriptions of the constituent CVEs as well as comments from security experts on the notification in the organization. There are a small percentage of notifications that do not comprise of any CVEs, and are descriptions of possible nonstandard vulnerability issues for assets.

An evaluation is an assessment of how a notification affects an asset. When a notification is known to affect a component that constitutes an asset, an evaluation is performed by a security expert who has thorough knowledge of the asset. The Evaluations dataset consists of the asset id, notification id, \VexCategory, \VexJustification, Internal comment, Customer Comment, Vector and CVSS Score. \VexCategory represents expert assessment of whether the asset is affected by the notification or not. Internal comment is the comment outlining mitigation process if the product is affected or short description of why the product is not affected. Customer Comment is a summarised version of Internal Comment which is sent to customers of the organization's products. \VexJustification comes into purview only when \VexCategory is `Not Affected', and is a justification of why the product is not affected. It can be one of the following categories:
\begin{enumerate}
    \item VulnerableCodeNotPresent
    \item ComponentNotPresent
    \item VulnerableCodeNotInExecutePath
    \item VulnerableCodeCannotBeControlledByAdversary
    \item InlineMitigationsAlreadyExist, and
    \item Other
\end{enumerate}
Vector corresponds to the evaluated CVSS vector which consists of base vector, temporal vector and environmental vector. CVSS score is calculated using the CVSS vectors.

There are around 1.6K assets, 134K notifications and 174K evaluations at the time of writing this paper. There are 152K unique components used by all the assets. There are 11K unique notifications that are used by all the evaluations. Most notification descriptions consist of less than 200 words, most internal comments have less than 70 words and most customer comments are less than 50 words \ref{fig:data_stats}. The instruction tuning dataset is formed from the Evaluations dataset. An instruction in the dataset corresponds to an evaluation and contains description of asset, description of notification, the affected software components, and the corresponding assessments: \VexCategory, \VexJustification, Internal Comment, Customer Comment, and Vector.

\subsection{Training of DAPT model}
DAPT model is trained by continuously pretraining the LLM using next token generation. Since the quality of data is of utmost importance, we proceed with careful preprocessing of DAPT dataset, and then pretraining the LLM with preprocessed data.

\subsubsection{Preprocessing}
Preprocessing of the public dataset part of DAPT dataset consists of the following steps:
\begin{itemize}
    \item Identification of attributes of the dataset that are used to form a data point. A data point in the dataset refers to details of a CVE. The identified attributes represent title and descriptions of the CVEs, base and temporal vectors, affected product and software version, unaffected software version (if any), and mitigation (if any).
    \item Cleaning descriptions. The descriptions are preprocessed for removing URLs, non-UTF8 characters, removing formatting if any. In case there are more than one description in the CVE, and there is not more than 70\% overlap between the descriptions, they are merged into one longer description.
    \item Template-based formation of vector description. The vectors are expanded to form template-based description of the vectors. For a vector component that is of the format <VECTOR\_COMPONENT:value>, the expanded vector description is ``<VECTOR COMPONENT> is <value>". 
    \item Template-based combination of the data point attributes.The template for the CVE detail is:
    ``CVE description: <description>. Affected product: <product> less than <highest affected software version>. Unaffected version:  <lowest unaffected version> and higher. Vector: <vector description>"
\end{itemize}

Preprocessing of the organization dataset consists of the following procedures:

\begin{itemize}
    \item Identification of notifications in the Notifications dataset that consist of descriptions with text other than the descriptions of the constituent CVEs. Most notifications contain CVE descriptions from the public dataset followed by expert comments on the notification. In addition, the nonstandard notifications are also included in this dataset.
    \item Cleaning description. This follows the same procedure of cleaning descriptions as the public dataset preprocessing. In addition, the CVE ids are cleaned from the description. 
    \item Template-based formation of vector description. Vector description is formed in the similar fashion as the public dataset.
    \item Template-based combination of notification attributes. The template for notification detail is:
    ``Notification description: <description>. Vector: <vector description>"
\end{itemize}

\subsubsection{Model Training}
The DAPT dataset is randomly split into 90:10 split with 288K training data points and 32K validation data points. We have continuously pretrained MPT-7B \cite{MosaicML2023Introducing} base model autoregressively for next token prediction with an objective function of cross entropy with the DAPT dataset. We have expanded the vocabulary of the MPT-7B base model to include names of the components and the organization software. We have used only the first names of the components and software and we had 539 new tokens to add to the 50K vocabulary of MPT. The model is trained using DeepSpeed \cite{rasley2020deepspeed} zero-3 optimization with Lion optimizer \cite{chen2024symbolic} (learning rate: 1e-4, weight decay: 1e-2) for 3 epochs on 8 Tesla A100 SXM4 GPUs.

\subsection{Training of Instruction-tuned model}

    \begin{table*}
      \caption{Instructions for each evaluation type}
      \label{tab:instr}
      \begin{tabular}{ccl}
        \toprule
        Evaluation&Instruction&Response\\
        \midrule
        \VexCategory/\VexJustification & What is the category?& <\VexJustification> Category: <VexCategory>\\
        Internal Comment & Generate internal comment.& <InternalComment>\\
        Customer Comment & Generate customer comment. & <CustomerComment>\\
        Vector & Generate environmental vectors. & <Evaluation vector description>\\
      \bottomrule
    \end{tabular}
    \end{table*}
    
Instruction tuning requires formation of an instruction tuning dataset in a pre-defined format. We have used the instruction tuning dataset format of the Alpaca dataset \cite{taori2023alpaca} to form our dataset. Then we have finetuned the DAPT model for instruction tuning with the new dataset.

\subsubsection{Formation of Instruction-tuning dataset}
Formation of instruction-tuning dataset follows the following procedure:
\begin{itemize}
    \item Merging of Evaluations dataset with Assets and Notifications dataset based on the Asset ID and Notification ID.
    \item Cleaning of notification description. This follows similar procedure as preprocessing of organization dataset.
    \item Identifying components common to both Asset and Notification and forming a list of descriptions of these common components.
    \item Formation of vector descriptions for both notification vector and evaluation vector. This follows the same vector text description format as described in DAPT dataset formation section. For evaluation vector description formation, we subtract the components already present in the notification vector from evaluation vector components and form the evaluation vector description using these new vector components only. In addition, the CVSS format version is also extracted. In case CVSS format version is not specified, the vector components are used to match to the corresponding vector components for CVSS versions 2, 3.0 and 3.1, and corresponding the highest version with all the component matches is considered as the CVSS version. In case such version is not found, we have used the help of security experts to correct the vector as well as get the correct version number. 
    
    We obtain all the expected base, temporal and environmental vector components using the CVSS version. In case of notification vector text, we find all missing base and temporal vector components in the text and add them to notification vector description in the format ``<VECTOR\_COMPONENT> is XXXX". For evaluation vector, we find all missing environmental vector components in the evaluation vector and add them to the evaluation vector text in the same format as missing notification vector description. To be noted is that the order of vectors in the description is fixed.
    \item Formation of vector justification. We break up the vector justification category into constituent words and use that as an explanation of \VexCategory.
    \item Formation of instruction data. The instruction format uses the following format:

    \begin{table}[!t]
    \begin{center}
    \begin{small}
    \begin{tabular}{m{0.98\linewidth}}
    \hline
    \multicolumn{1}{c}{Instruction dataset format}\\
    \hline
    Below is an instruction that describes a task, paired with an input that provides further context.
    Write a response that appropriately completes the request.

    \#\#\# Instruction: <instruction>
    
    \#\#\# Input:
    
    Organization: <sub-organization name>
    
    Software: <software name and version>
    
    Product: <asset name and version>
    
    Notification: <cleaned notification description>
    
    Components present in software: <list of common component description>
    
    Base and Temporal Vectors: <notification vector description>

    CVSS Version: <version>

    \#\#\# Response: <response>
    
    <STOP>
    \\
    \hline
\multicolumn{1}{c}{Example}\\
\hline
    Below is an instruction that describes a task, paired with an input that provides further context.
    Write a response that appropriately completes the request.

    \#\#\# Instruction: Generate internal comments.
    
    \#\#\# Input:
    
    Organization: \censor{DI-DnA}
    
    Software: \censor{Syngo Carbon Monitoring}
    
    Product: \censor{Syngo Carbon Monitoring VB12A}
    
    Notification: 
    
    CoreMedia, WebRTC: Processing an image can cause an out-of-bounds write and may lead to arbitrary code execution due to improper input validation.
    
    Components present in software: Debian Package: dav1d - Debian - All Versions
    
    Base and Temporal Vectors: Attack Vector is Adjacent. Attack Complexity is High. Privileges Required is Low. User Interaction is None. Scope is Unchanged. Confidentiality is Low. Integrity is High. Availability is Low. Exploit Code Maturity is Unproven. Remediation Level is Official Fix. Report Confidence is Confirmed.

    CVSS Version: 3.1

    \#\#\# Response: CoreMedia, WebRTC is not utilized in the product.
    
    <STOP>
    \\
    \hline
    \hline
    \end{tabular}
    \end{small}
    \end{center}
    \caption{
    Instruction tuning dataset format and example.\\
    Certain data is censored in keeping with organization policies.} 
    \end{table}

    The instructions and responses for the different types of evaluations are shown in \ref{tab:instr}. 

    \item Removal of incomplete evaluation data, data under investigation, long text data. In practice, \VexCategory also has two more classes: UnderInvestigation and EndofLife. UnderInvestigation category is assigned to the evaluations for which monitoring and/or mitigation measures are underway. EndofLife refers to the assets that are being retired. These categories are flagged by the system and we do not use the data to train our model. In addition, there are a few data points identified that had incomplete evaluations. We also identified a few data points for which the token count exceeds 1048 tokens, and we do not use that data as well for training.

    \item De-duplication of dataset. The dataset is de-duplicated by removing the data points for which all software name, notification description and evaluations are the same. 
    
\end{itemize}

 \subsubsection{Model Training}
 The instruction tuning dataset has 440K instructions, and the dataset is randomly split into train, test and validation datasets with a 80:10:10 split, leading to 352k training data points, and 44K each of validation and test data points. The DAPT model is finetuned with the training and validation data by back propagating on completions only, with setting the token labels of the instructions to -100. The model is trained using DeepSpeed zero-3 optimization with Lion optimizer (learning rate: 1e-5, weight decay: 1e-2) for 3 epochs on 8 Tesla A100 SXM4 GPUs.

 \subsection{Inference}


 \begin{figure*}
  \centering
  \begin{frame}{\includegraphics[width=\linewidth]{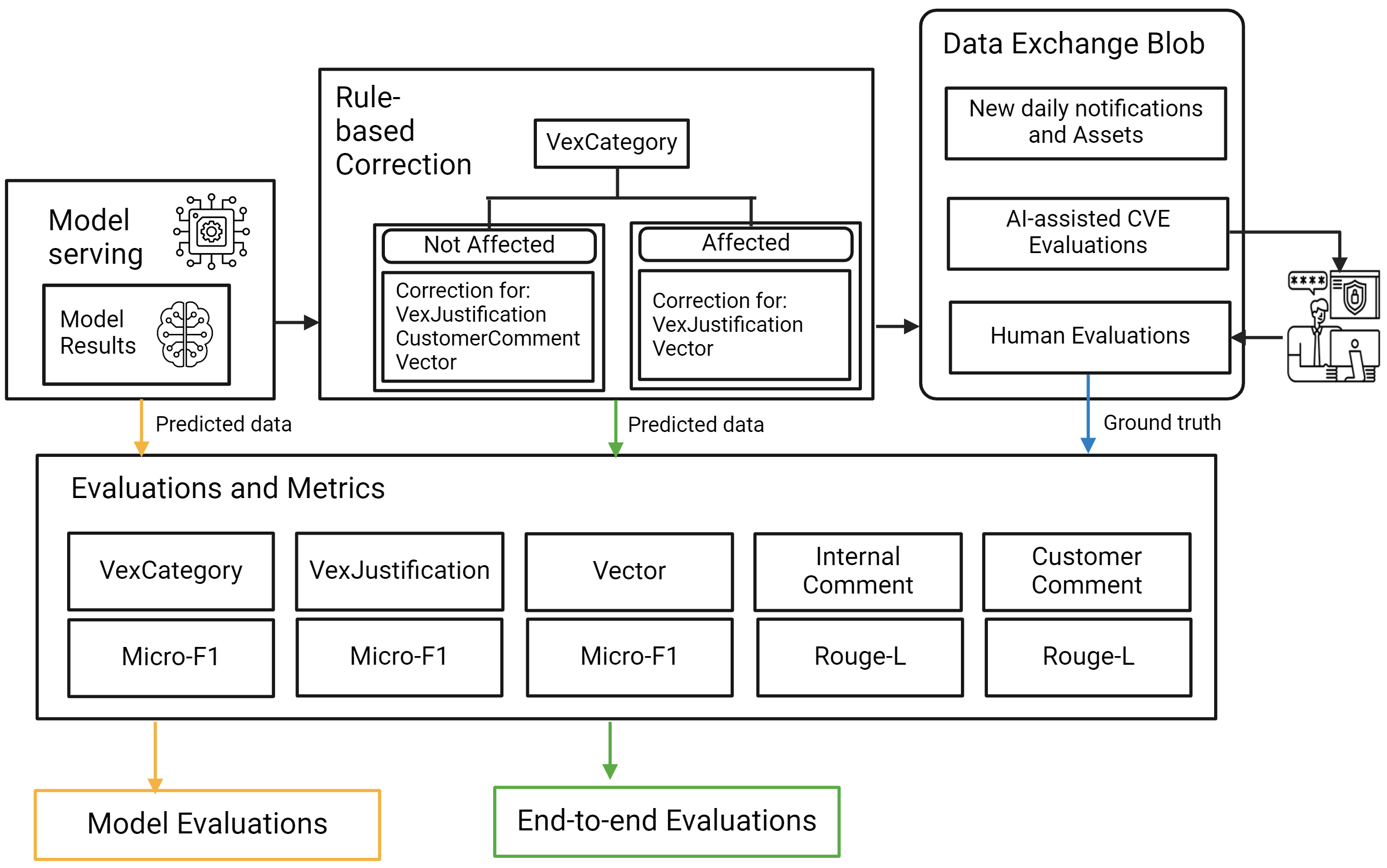}}
      
  \end{frame}
  \caption{Inference and Model evaluation Schematic}
  \Description{Inference Schema}
\end{figure*}

 The inference framework includes generation of evaluations using the trained LLM, using pre-defined rules to avoid easily identifiable erroneous generations, and using faster inference frameworks to improve model throughput.  We have implemented a human-in-the-loop system where the AI-generated vulnerability evaluations are validated by a product cybersecurity expert with higher priority to the evaluations with Affected \VexCategory.

 \subsubsection{Generation of evaluations}
 Evaluation generation follows a zero shot model inference where the model is provided a prompt in the pre-defined instruction format, containing information about sub-organization name, software name and version, asset name and version, notification description, list of component descriptions for components present in both asset and notification, base and temporal versions and CVSS version. The maximum generated tokens for \VexCategory/\VexJustification generation is set to 25, Internal Comments is 125, Customer comments is 100, and Vectors is 100. The inference/test dataset is divided into two sub-datasets based on the size of tokens in the inference instruction. For instructions longer than 920 tokens, the sequence length of the trained model is increased to 150 + maximum token length of the longest instruction. If the size of the instruction is less than 920 tokens, the trained model without any changes is used to inference the dataset. Since MPT-7B is trained with ALIBI \cite{press2021train} encodings, the maximum sequence length during inference can be set upto 84K tokens, which has not been observed in production yet.

 \subsubsection{Rule-based correction}
 Post-processing of model outputs involves using system knowledge in the form of pre-defined rules to correct mistakes in model generation. The rules are as follows: 
 \begin{enumerate}
    \item If \VexCategory value is NotAffected, the evaluation vector is not generated.
    \item If \VexJustification generated is any text other than the pre-defined categories, `Other' gets assigned to \VexJustification.
    \item If \VexJustification is `Other' and Customer Comment is empty, then it is an error in generation. We use Internal Comment as External Comment as well in case it is not empty.
    \item If \VexCategory is Affected, \VexJustification is automatically set to None, and it is ensured that the Vector has the proper CVSS format.

 \end{enumerate}

 \subsubsection{Model serving}
 Model serving is performed using vLLM architecture \cite{kwon2023efficient} on one Tesla A100 SXM4 GPU. Model throughput during production is improved using rule based system where generation of Vector is skipped when \VexCategory is Not Affected, and sequence length-based batching is performed.

\section{Results}
Evaluation of our methodology follows a two-step process which evaluate the model-generated results as well as the end-to-end results from the inference pipeline. In addition to the test set which was set apart during training, we have also conducted an evaluation of the model outputs in a real-time deployment scenario over a period of two months for new assets and vulnerabilities.
\subsection{Evaluation Metrics}
We use ROUGE-L and micro-F1 for evaluating the model responses: ROUGE-L\cite{lin2004looking} is used for evaluating the responses model generated for Internal Comment and Customer Comment, whereas micro-F1\cite{scikit-learn} is used for \VexCategory, \VexJustification and Vector. ROUGE-L (Recall-Oriented Understudy for Gisting Evaluation - Longest Common Subsequence) is a metric used to evaluate the quality of summaries produced by automatic summarization systems. ROUGE-L measures the longest common subsequence (LCS) between the model-generated summary and the reference (human-generated) summary. Micro F1, or micro-averaged F1 score, is calculated by taking the harmonic mean of precision and recall, with equal weighting given to both metrics. 
\subsection{Benchmarking CVE-LLM against other open source LLMs}
We have used the open source models that are of similar size as CVE-LLM, and are the highest performing models currently on LLM Leaderboards \cite{llm_leaderboard}. We trained them using the DAPT and SFT methods described in Methodology section. In addition, we have also performed RAG over our test dataset with Llama3-70B. The Table \ref{tab:llmtest} enumerates the results on our test datasets for these experiments, as well as the performance of our system in production. Except CVE-LLM-Prod, which is the system deployment results against 2K expert evaluations over completely new notifications and assets, the rest are evaluated on the test dataset with 44K evaluations.

\begin{table*}
    \begin{tabular}{ccccccl}
        \toprule
        Evaluation&Llama3-RAG&Mistral-7B&Llama2-7B&CVE-LLM-Eval&CVE-LLM-Eval*&CVE-LLM-Prod\\
        \midrule
        \VexCategory & 0.25& 0.64& 0.56&0.93&0.93&0.83\\
        \VexJustification & 0.12&0.6& 0.53&0.89&0.92&0.95\\
        Internal Comment & 0.19&0.71& 0.62&0.8&0.8&0.7\\
        Customer Comment & 0.28&0.7 & 0.64&0.88&0.89&0.75\\
        Vector & 0.13&0.59 & 0.57&0.95&0.97&0.97\\
      \bottomrule
    \end{tabular}
    \caption{Benchmarking CVE-LLM against state-of-the-art Open Source LLMs \\CVE-LLM-Eval* = End-to-end CVE-LLM results, CVE-LLM-Prod = CVE-LLM-Production results.}
    \label{tab:llmtest}
    \end{table*}
CVE-LLM, which is based on MPT-7B model, has shown better performance than Llama2-7B and Mistral-7B-based models. The highest performance improvement is observed for classification-based generations: \VexCategory, \VexJustification and Vector. Post processing the model results have shown the highest improvement for \VexJustification and Vector. We also observe that the performance of CVE-LLM with completely new assets and notifications in deployment mirrors the test dataset results. 

\subsection{Ablation Studies with the model}

We conducted ablation studies with respect to different components of model training and inference: 

\begin{inparaenum}[(a)]
    \item Changes in dataset,
    \item Dataset size,
    \item DAPT-SFT vs SFT-only system, and
    \item Impact of beam size, temperature, and nucleus sampling during model inference, and
    \item Sequence length during training, and
\end{inparaenum}
For each of these experiments, the test dataset consists of the same data points from the Evaluations dataset. The training data points have varied across the datasets, but we have made sure that all the possible assets are represented in the training datasets.

\subsubsection{Changes in dataset}

\begin{table}
      \begin{tabular}{ccccccl}
        \toprule
         Evaluation&CI&CV&CWJ&WD&NP\\
        \midrule
        \VexCategory &0.87&0.94&0.85&0.66&0.93\\
        \VexJustification &0.86&0.89&0.85&0.51&0.89\\
        Internal Comment &0.7&0.79&0.8&0.66&0.8\\
        Customer Comment &0.79&0.81&0.88&0.62&0.88\\
        Vector &0.89&0.9&0.95&0.7&0.85\\
      \bottomrule
    \end{tabular}
    \caption{CVE-LLM-Eval Model Ablation study on Test Dataset\\CI = All Instructions Combined, CV = Customer Comment and \VexCategory/\VexJustification combined, CWJ = \VexCategory and \VexJustification separated, SL = Sequence length, WD=Without delimiter, NP = No Placeholder for unused Vector Categories}
    \label{tab:testablation}
\end{table}

For each of these experiments, we trained CVE-LLM-base model on the reconstructed dataset. We have tried the following changes in dataset:
\begin{itemize}
    \item Combining all the evaluations into one instruction (CI)
    \item Combining customer comments, \VexCategory and \VexJustification into one type of instruction (CV)
    \item Separate generation of \VexJustification and \VexCategory (CWJ)
    \item No delimiter at the end of training text (WD), and
    \item No Placeholder for unused vector categories (NP)
\end{itemize}
CI, CV and CWJ are concerned with experiments to see whether we can combine some of these instructions to achieve good results. WD refers to the experiment where we see how the model performs when we do not put a dedicated delimiter. NP is concerned with how the template for a Vector output should look like. NP refers to the case where the response contains only the vectors that are mentioned in the standard CVSS vector format. The training details of this ablation study is shown in Table \ref{tab:traindeets} and the test results are shown in Table \ref{tab:testablation}. These datasets vary in their sizes of their training, validation and test datasets. For example, CI has only 88K instructions to train with due to combining instructions, which CWJ has 440K instructions because all the instructions are separated.

\begin{table}
      \begin{tabular}{cccccl}
        \toprule
         &CI&CV&CWJ&WD\\
        \midrule
        $N_{Train}$ &88K&264K&440K&329K\\
        $N_{Valid}$ &11K&33K&55K&44K\\
        $N_{Test}$ &11K&33K&55K&44K\\
        Epochs &4&2&3&3\\
      \bottomrule
    \end{tabular}
    \caption{Changes in Dataset: CVE-LLM-Eval Model Ablation study on Test Dataset\\CI = All Instructions Combined, CV = Customer Comment and \VexCategory/\VexJustification combined, CWJ = \VexCategory and \VexJustification separated, WD=Without delimiter}
    \label{tab:traindeets}
\end{table}

We observe that the variation of Rouge-L score for Internal Comment is the lowest, while the highest variation is seen for the classification-based generations of \VexCategory and Vector. The model performs better with diversity in instructions, and higher number of tokens in training. Size of the dataset does not impact performance much by itself, but combined with other factors may degrade performance. Spurious text generation is a problem with LLM finetuning, and putting a delimiter enhances performance more than changing model generation parameters. Vector performance improves by 0.1 micro-F1 with the introduction of fixed template-based generation for all possible components of the environmental and temporal vector. 

\subsubsection{Effect of Dataset size}
We have experimented with changing size of the instruction tuning training dataset from 150K instructions to 440K instructions. The datasets are formed by making sure that all the types of assets and all the types of notifications are always represented in the dataset. The minimal dataset that affords this invariability is the 150K training dataset, and for the rest of the datasets we have randomly selected from the remaining data points.

\begin{figure*}
    \centering
    \begin{subfigure}[b]{0.3\textwidth}
        \includegraphics[width=\textwidth]{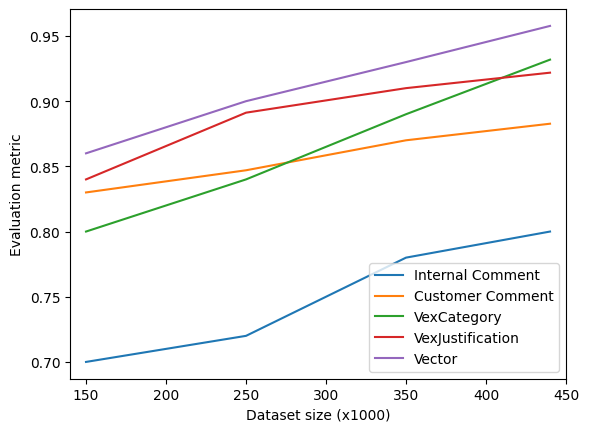}
        \caption{Effect of dataset size}
        \label{fig:dsize}
    \end{subfigure}
    \begin{subfigure}[b]{0.3\textwidth}
        \includegraphics[width=\textwidth]{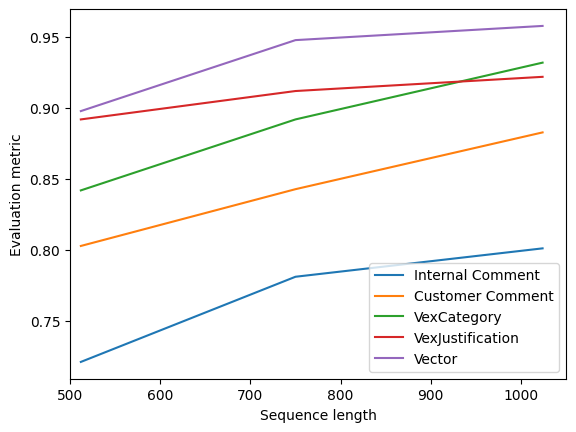}
        \caption{Effect of sequence length}
        \label{fig:seq_len}
    \end{subfigure}
    \begin{subfigure}[b]{0.3\textwidth}
        \includegraphics[width=\textwidth]{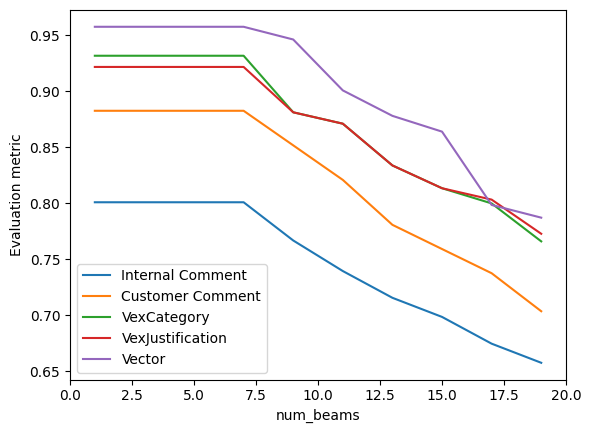}
        \caption{Effect of beam size}
        \label{fig:beam}
    \end{subfigure}
    \caption{Ablation studies: Effect of various factors on model outcome}\label{fig:ablation}
\end{figure*}

We have noticed that Customer Comment shows the least variation in Rouge-L score, while \VexCategory shows the highest micro-F1 variance. In addition, the performance improvement stagnates for most instructions as we reach the 440K instruction dataset size mark.

\subsubsection{Effect of Domain Adaptation}

The results after Supervised Finetuning of Domain-adapted CVE-LLM-Base vs Supervised Finetuning of MPT-7B is shown in Table \ref{tab:testdapt}. Domain adaptation and vocabulary expansion leads to a much better performance for all the instruction types, as shown in Table \ref{tab:testdapt}
\begin{table}
      \begin{tabular}{ccl}
        \toprule
        Evaluation&CVE-LLM-Base&MPT-7B\\
        \midrule
        \VexCategory & 0.93&0.83\\
        \VexJustification & 0.89& 0.8\\
        Internal Comment & 0.8&0.64 \\
        Customer Comment & 0.88 & 0.69 \\
        Vector & 0.95 & 0.84\\
      \bottomrule
    \end{tabular}
    \caption{Test dataset Results: Finetuning Domain adapted LLM CVE-LLM-Base vs Finetuning with MPT-7B}
      \label{tab:testdapt}
    \end{table}

\subsubsection{Effect of Sequence Length in Training}
In this experiment, we changed the sequence length in configuration of the model to 512, 750 and 1024, and reduced the training dataset to accommodate only the instructions that had less than the sequence length. Figure \ref{fig:seq_len} shows the effect of training sequence length on the model outcome for all the instruction types. We find out that though the dataset size does not change much, there is more variation in performance. As training sequence length increases, the performance on datasets with longer instructions also improves.

\subsubsection{Effect of inference parameters}
In generative model inference, parameters such as temperature, beam size, and nucleus sampling significantly influence the quality and diversity of the generated outputs. Temperature controls the randomness of the output, with higher values leading to more diverse results. Nucleus Sampling (top-p sampling) dynamically adjusts the candidate set of tokens to balance between diversity and coherence based on the cumulative probability threshold `p'. Beam Size in beam search influences the depth of search for probable sequences, balancing between computational cost and output quality. Generally these parameters are adjusted to optimize the trade-off  between diversity and coherence of the generated text. We performed inference on test dataset (N=44K) using CVE-LLM-Eval model to conduct the experiment on the effect of inference parameters.

\paragraph{Temperature and nucleus sampling.} In order to test the effect of temperature, we maintained the generation parameters at their default settings and conducted tests by varying the temperature values from 0 to 1 in increments of 0.1 to evaluate the outcomes. Similarly, for top-p sampling too, we changed the parameter `top-p' from 0 to 1 in increments of 0.1, keeping all other generation parameters in default settings. For both of the experiments, no changes were observed.

\paragraph{Beam size.} We measured the effect of beam size for each of the instruction types by changing the beam size from 1 to 19 in increments of 2, maintaining other generation parameters in their default values. We notice that beyond beam size 7, the performance of the system steadily declines for all the instruction types. The effect of change of beam size on model generation is shown in Figure \ref{fig:beam}.

\subsection{Inference time}
We have used different techniques to optimize for the time required in inference. The first experiment used popular model serving algorithm vLLM adapted for MPT-7B to find required inference time for one Tesla A100 SXM4 GPU for each of the instruction types and found speedup of ten times using both compared to no model serving. The experiment results are shown in Table \ref{tab:testinf}

\begin{table}
      \begin{tabular}{cccl}
        \toprule
        Evaluation&No speedup&vLLM\\
        \midrule
        \VexCategory/\VexJustification & 0.5&0.04\\
        Internal Comment & 2.5&0.2 \\
        Customer Comment & 2& 0.19 \\
        Vector & 3&0.32\\
      \bottomrule
    \end{tabular}
    \caption{Inference time in seconds on 44K test dataset for Model serving with sequence length = 20K}
      \label{tab:testinf}
    \end{table}
In addition, the adaptive generation by dividing the inference batch into batches of small and large token lengths also improves the average performance over the test dataset to 4.5 second per evaluation from 8 second. The other adaptation of using different maximum sequence length for each instruction type brings down the inference time for each data point in the Evaluations dataset by 1.5 times.

We have calculated the time required by a cybersecurity expert in our organization to evaluate vulnerabilities. The time required for an expert, averaged over 500 evaluations is 229s, whereas an evaluation by an our model takes only 1 second per evaluation.

\section{Discussion}
In this section, we will elucidate our observations related to training and performance of LLMs for vulnerability management. We have learned valuable lessons in understanding the benefits and pitfalls of using LLMs, and have identified areas of improvement for the system.

\subsection{Observations on Model Training}
In our experiments we tested several hypothesis about diversity of instructions and dataset size. Firstly, we observe that the size of the dataset does not independently impact performance; rather, the diversity of instructions significantly influences the model's efficacy. Higher diversity leads to better performance. When all instructions were consolidated into a single instruction type, there was a marked decline in performance. This decrease was primarily due to the model generating extraneous text, either supplementing or replacing the expected response. Secondly, the responses corresponding to instruction types that appeared later in the consolidated instruction model had higher degradation of performance. This is corroborated by training with combination of customer comments and category as well. Thirdly, representation of related assets and notifications are important, since the model performance dips when tested on completely different types of assets and notifications.

LLMs have reportedly underperformed in text classification tasks; however, their performance has improved when reasoning elements are incorporated into the text \cite{sun2023text}. Our experiments yielded similar observations. We experimented on using reasoning for \VexCategory and found that this leads to higher F1 than without using reasoning. In our study, we used \VexJustification as a reasoning for \VexCategory, as well as examined whether adding Customer Comment as an explanation would improve the system. There are no major improvements with adding Customer Comments as reasoning and also leads it to lower performance for Customer Comments. Chain-of-Thought prompts which have been shown to enhance model reasoning power would probably be useful to enhance our model in the future. Additionally, we have observed that the total number of classes affect the performance of the system greatly. Vector generation can be considered as a multi-label multiclass classification problem whereas \VexCategory is a binary classification problem and \VexJustification is a multiclass classifcation problem. Though the F1 for Vector is high, F1 is still low for underrepresented classes which leads to decrease in performance. The performance also drops from \VexCategory classification to \VexJustification classification, thereby showing that the number of classes affects the performance. However, the Customer Comment section follows a highly templated format, with approximately 85\% of customer comments being repeated across multiple evaluations. Hence we might consider Customer Comment also to be a classification problem with large number of classes. However the performance on Customer Comments does not stray far from the performance for \VexJustification with much fewer classes.

\subsection{Observations on Inference}
We have noticed model output variations with beam size, with decline in performance with higher beam sizes. Increase in beam search size increases the diversity of text, and that leads to less contextually relevant outputs, as seen in our experiments. Low beam size is sufficient for high quality generation because of the specialized nature of the dataset. The difference between probabilities of less probable outcomes and highly probable outcomes, given the context, is low. This leads to no temperature or nucleus sampling-based variations in model-generated outputs.

\subsection{Error analysis}
We have used MPT-7B to train CVE-LLM for two reasons:
\begin{inparaenum}
    \item Expansion capability to at least 20K tokens without significant loss of performance, and
    \item It performs better than other open source models we tested with our dataset
\end{inparaenum}
However, among all the LLMs we have trained, majority of the issues we have encountered are remarkably similar, differing primarily in their severity. The generated text for both Internal Comment and Customer Comment has been observed to have omitted or falsely included several critical details, particularly those concerning the affected software versions, the recommended update versions, the names of the software or components, and other similarly named entities that constitute essential information. The other type of error commonly seen for the LLMs is spurious text generation, which is something we observed quite frequently for trained Llama2 models. Introduction of stop tokens mitigated the problem to a great extent for MPT-7B, but not for the other models. In spite of having the ability to extend the sequence length of the trained model beyond the training sequence length, the performance on long sequences still falls behind. Long instructions comprises of multiple vulnerabilities, and constitute only 10\% of our test dataset. In the future, we plan to segment it into multiple vulnerabilities and use more robust semantic understanding of each vulnerability to perform vulnerability chaining. We also tested the system only with zero shot instructions at the inference, and we surmise that we can utilize language understanding capabilities better by using few shot methods to assess vulnerabilities for assets.


This study illustrates the capacity of large language models (LLMs) to learn from expert-curated historical data, thereby enabling the automation of vulnerability management for medical devices. Our model CVE-LLM has proved to be effective in learning from past evaluations and has shown high accuracy in prediction of \VexCategory and CVSS Vectors. The model's inference is also substantially faster, achieving speeds approximately 200 times greater than those of a human expert. We intend for our system to be employed in a human-in-the-loop manner, assisting resident product cybersecurity experts in swiftly identifying vulnerabilities that affect the assets and communicating promptly to the customer.

\section{Disclaimer}
The concepts and information presented in this paper/presentation are based on research results that are not commercially available. Future commercial availability cannot be guaranteed.


\bibliographystyle{ACM-Reference-Format}
\bibliography{main}


\begin{thebibliography}{68}


\ifx \showCODEN    \undefined \def \showCODEN     #1{\unskip}     \fi
\ifx \showDOI      \undefined \def \showDOI       #1{#1}\fi
\ifx \showISBNx    \undefined \def \showISBNx     #1{\unskip}     \fi
\ifx \showISBNxiii \undefined \def \showISBNxiii  #1{\unskip}     \fi
\ifx \showISSN     \undefined \def \showISSN      #1{\unskip}     \fi
\ifx \showLCCN     \undefined \def \showLCCN      #1{\unskip}     \fi
\ifx \shownote     \undefined \def \shownote      #1{#1}          \fi
\ifx \showarticletitle \undefined \def \showarticletitle #1{#1}   \fi
\ifx \showURL      \undefined \def \showURL       {\relax}        \fi
\providecommand\bibfield[2]{#2}
\providecommand\bibinfo[2]{#2}
\providecommand\natexlab[1]{#1}
\providecommand\showeprint[2][]{arXiv:#2}

\bibitem[Abdeen et~al\mbox{.}(2023)]%
        {abdeen2023smet}
\bibfield{author}{\bibinfo{person}{Basel Abdeen}, \bibinfo{person}{Ehab Al-Shaer}, \bibinfo{person}{Anoop Singhal}, \bibinfo{person}{Latifur Khan}, {and} \bibinfo{person}{Kevin Hamlen}.} \bibinfo{year}{2023}\natexlab{}.
\newblock \showarticletitle{Smet: Semantic mapping of cve to att\&ck and its application to cybersecurity}. In \bibinfo{booktitle}{\emph{IFIP Annual Conference on Data and Applications Security and Privacy}}. Springer, \bibinfo{pages}{243--260}.
\newblock


\bibitem[Achiam et~al\mbox{.}(2023)]%
        {achiam2023gpt}
\bibfield{author}{\bibinfo{person}{Josh Achiam}, \bibinfo{person}{Steven Adler}, \bibinfo{person}{Sandhini Agarwal}, \bibinfo{person}{Lama Ahmad}, \bibinfo{person}{Ilge Akkaya}, \bibinfo{person}{Florencia~Leoni Aleman}, \bibinfo{person}{Diogo Almeida}, \bibinfo{person}{Janko Altenschmidt}, \bibinfo{person}{Sam Altman}, \bibinfo{person}{Shyamal Anadkat}, {et~al\mbox{.}}} \bibinfo{year}{2023}\natexlab{}.
\newblock \showarticletitle{Gpt-4 technical report}.
\newblock \bibinfo{journal}{\emph{arXiv preprint arXiv:2303.08774}} (\bibinfo{year}{2023}).
\newblock


\bibitem[Ameri et~al\mbox{.}(2021)]%
        {ameri2021cybert}
\bibfield{author}{\bibinfo{person}{Kimia Ameri}, \bibinfo{person}{Michael Hempel}, \bibinfo{person}{Hamid Sharif}, \bibinfo{person}{Juan Lopez~Jr}, {and} \bibinfo{person}{Kalyan Perumalla}.} \bibinfo{year}{2021}\natexlab{}.
\newblock \showarticletitle{Cybert: Cybersecurity claim classification by fine-tuning the bert language model}.
\newblock \bibinfo{journal}{\emph{Journal of Cybersecurity and Privacy}} \bibinfo{volume}{1}, \bibinfo{number}{4} (\bibinfo{year}{2021}), \bibinfo{pages}{615--637}.
\newblock


\bibitem[Ampel et~al\mbox{.}(2021)]%
        {ampel2021linking}
\bibfield{author}{\bibinfo{person}{Benjamin Ampel}, \bibinfo{person}{Sagar Samtani}, \bibinfo{person}{Steven Ullman}, {and} \bibinfo{person}{Hsinchun Chen}.} \bibinfo{year}{2021}\natexlab{}.
\newblock \showarticletitle{Linking common vulnerabilities and exposures to the mitre att\&ck framework: A self-distillation approach}.
\newblock \bibinfo{journal}{\emph{arXiv preprint arXiv:2108.01696}} (\bibinfo{year}{2021}).
\newblock


\bibitem[Center(2024)]%
        {idtheftreport}
\bibfield{author}{\bibinfo{person}{Identity Theft~Resource Center}.} \bibinfo{year}{January 2024}\natexlab{}.
\newblock \bibinfo{booktitle}{\emph{2023 Data Breach Report}}.
\newblock
\urldef\tempurl%
\url{https://www.idtheftcenter.org/wp-content/uploads/2024/01/ITRC_2023-Annual-Data-Breach-Report.pdf}
\showURL{%
\tempurl}


\bibitem[Chen et~al\mbox{.}(2024)]%
        {chen2024symbolic}
\bibfield{author}{\bibinfo{person}{Xiangning Chen}, \bibinfo{person}{Chen Liang}, \bibinfo{person}{Da Huang}, \bibinfo{person}{Esteban Real}, \bibinfo{person}{Kaiyuan Wang}, \bibinfo{person}{Hieu Pham}, \bibinfo{person}{Xuanyi Dong}, \bibinfo{person}{Thang Luong}, \bibinfo{person}{Cho-Jui Hsieh}, \bibinfo{person}{Yifeng Lu}, {et~al\mbox{.}}} \bibinfo{year}{2024}\natexlab{}.
\newblock \showarticletitle{Symbolic discovery of optimization algorithms}.
\newblock \bibinfo{journal}{\emph{Advances in Neural Information Processing Systems}}  \bibinfo{volume}{36} (\bibinfo{year}{2024}).
\newblock


\bibitem[Christey et~al\mbox{.}(2013)]%
        {christey2013common}
\bibfield{author}{\bibinfo{person}{Steve Christey}, \bibinfo{person}{J Kenderdine}, \bibinfo{person}{J Mazella}, {and} \bibinfo{person}{B Miles}.} \bibinfo{year}{2013}\natexlab{}.
\newblock \showarticletitle{Common weakness enumeration}.
\newblock \bibinfo{journal}{\emph{Mitre Corporation}} (\bibinfo{year}{2013}).
\newblock


\bibitem[Chronopoulou et~al\mbox{.}(2019)]%
        {chronopoulou2019embarrassingly}
\bibfield{author}{\bibinfo{person}{Alexandra Chronopoulou}, \bibinfo{person}{Christos Baziotis}, {and} \bibinfo{person}{Alexandros Potamianos}.} \bibinfo{year}{2019}\natexlab{}.
\newblock \showarticletitle{An embarrassingly simple approach for transfer learning from pretrained language models}.
\newblock \bibinfo{journal}{\emph{arXiv preprint arXiv:1902.10547}} (\bibinfo{year}{2019}).
\newblock


\bibitem[Corporation(2024a)]%
        {attack_database}
\bibfield{author}{\bibinfo{person}{The~MITRE Corporation}.} \bibinfo{year}{2024}\natexlab{a}.
\newblock \bibinfo{booktitle}{\emph{MITRE ATT\&CK}}.
\newblock
\urldef\tempurl%
\url{https://attack.mitre.org/}
\showURL{%
\tempurl}


\bibitem[Corporation(2024b)]%
        {capec_database}
\bibfield{author}{\bibinfo{person}{The~MITRE Corporation}.} \bibinfo{year}{2024}\natexlab{b}.
\newblock \bibinfo{booktitle}{\emph{MITRE CAPEC}}.
\newblock
\urldef\tempurl%
\url{https://capec.mitre.org/}
\showURL{%
\tempurl}


\bibitem[Devlin et~al\mbox{.}(2018)]%
        {devlin2018bert}
\bibfield{author}{\bibinfo{person}{Jacob Devlin}, \bibinfo{person}{Ming-Wei Chang}, \bibinfo{person}{Kenton Lee}, {and} \bibinfo{person}{Kristina Toutanova}.} \bibinfo{year}{2018}\natexlab{}.
\newblock \showarticletitle{Bert: Pre-training of deep bidirectional transformers for language understanding}.
\newblock \bibinfo{journal}{\emph{arXiv preprint arXiv:1810.04805}} (\bibinfo{year}{2018}).
\newblock


\bibitem[Elbaz et~al\mbox{.}(2020)]%
        {elbaz2020fighting}
\bibfield{author}{\bibinfo{person}{Cl{\'e}ment Elbaz}, \bibinfo{person}{Louis Rilling}, {and} \bibinfo{person}{Christine Morin}.} \bibinfo{year}{2020}\natexlab{}.
\newblock \showarticletitle{Fighting N-day vulnerabilities with automated CVSS vector prediction at disclosure}. In \bibinfo{booktitle}{\emph{Proceedings of the 15th International Conference on Availability, Reliability and Security}}. \bibinfo{pages}{1--10}.
\newblock


\bibitem[Feng et~al\mbox{.}(2023)]%
        {feng2023knowledge}
\bibfield{author}{\bibinfo{person}{Chao Feng}, \bibinfo{person}{Xinyu Zhang}, {and} \bibinfo{person}{Zichu Fei}.} \bibinfo{year}{2023}\natexlab{}.
\newblock \showarticletitle{Knowledge solver: Teaching llms to search for domain knowledge from knowledge graphs}.
\newblock \bibinfo{journal}{\emph{arXiv preprint arXiv:2309.03118}} (\bibinfo{year}{2023}).
\newblock


\bibitem[Fu et~al\mbox{.}(2024)]%
        {fu2024constrained}
\bibfield{author}{\bibinfo{person}{Yanjun Fu}, \bibinfo{person}{Ethan Baker}, {and} \bibinfo{person}{Yizheng Chen}.} \bibinfo{year}{2024}\natexlab{}.
\newblock \showarticletitle{Constrained Decoding for Secure Code Generation}.
\newblock \bibinfo{journal}{\emph{arXiv preprint arXiv:2405.00218}} (\bibinfo{year}{2024}).
\newblock


\bibitem[Ge et~al\mbox{.}(2024)]%
        {ge2024openagi}
\bibfield{author}{\bibinfo{person}{Yingqiang Ge}, \bibinfo{person}{Wenyue Hua}, \bibinfo{person}{Kai Mei}, \bibinfo{person}{Juntao Tan}, \bibinfo{person}{Shuyuan Xu}, \bibinfo{person}{Zelong Li}, \bibinfo{person}{Yongfeng Zhang}, {et~al\mbox{.}}} \bibinfo{year}{2024}\natexlab{}.
\newblock \showarticletitle{Openagi: When llm meets domain experts}.
\newblock \bibinfo{journal}{\emph{Advances in Neural Information Processing Systems}}  \bibinfo{volume}{36} (\bibinfo{year}{2024}).
\newblock


\bibitem[Hassanin et~al\mbox{.}(2024)]%
        {hassanin2024pllm}
\bibfield{author}{\bibinfo{person}{Mohammed Hassanin}, \bibinfo{person}{Marwa Keshk}, \bibinfo{person}{Sara Salim}, \bibinfo{person}{Majid Alsubaie}, {and} \bibinfo{person}{Dharmendra Sharma}.} \bibinfo{year}{2024}\natexlab{}.
\newblock \showarticletitle{PLLM-CS: Pre-trained Large Language Model (LLM) for Cyber Threat Detection in Satellite Networks}.
\newblock \bibinfo{journal}{\emph{arXiv preprint arXiv:2405.05469}} (\bibinfo{year}{2024}).
\newblock


\bibitem[Huggingface(2023)]%
        {llm_leaderboard}
\bibfield{author}{\bibinfo{person}{Huggingface}.} \bibinfo{year}{2023}\natexlab{}.
\newblock \bibinfo{booktitle}{\emph{Open LLM Leaderboard}}.
\newblock
\urldef\tempurl%
\url{https://huggingface.co/spaces/HuggingFaceH4/open_llm_leaderboard}
\showURL{%
\tempurl}


\bibitem[Iorga et~al\mbox{.}(2020)]%
        {iorga2020early}
\bibfield{author}{\bibinfo{person}{Denis Iorga}, \bibinfo{person}{Dragos Corl{\u{a}}tescu}, \bibinfo{person}{Octavian Grigorescu}, \bibinfo{person}{Cristian S{\u{a}}ndescu}, \bibinfo{person}{Mihai Dasc{\u{a}}lu}, {and} \bibinfo{person}{Razvan Rughini{\c{s}}}.} \bibinfo{year}{2020}\natexlab{}.
\newblock \showarticletitle{Early detection of vulnerabilities from news websites using machine learning models}. In \bibinfo{booktitle}{\emph{2020 19th RoEduNet Conference: Networking in Education and Research (RoEduNet)}}. IEEE, \bibinfo{pages}{1--6}.
\newblock


\bibitem[Islam et~al\mbox{.}(2024)]%
        {islam2024llm}
\bibfield{author}{\bibinfo{person}{Nafis~Tanveer Islam}, \bibinfo{person}{Joseph Khoury}, \bibinfo{person}{Andrew Seong}, \bibinfo{person}{Gonzalo De La~Torre Parra}, \bibinfo{person}{Elias Bou-Harb}, {and} \bibinfo{person}{Peyman Najafirad}.} \bibinfo{year}{2024}\natexlab{}.
\newblock \showarticletitle{LLM-Powered Code Vulnerability Repair with Reinforcement Learning and Semantic Reward}.
\newblock \bibinfo{journal}{\emph{arXiv preprint arXiv:2401.03374}} (\bibinfo{year}{2024}).
\newblock


\bibitem[Jiang et~al\mbox{.}(2024)]%
        {jiang2024mixtral}
\bibfield{author}{\bibinfo{person}{Albert~Q Jiang}, \bibinfo{person}{Alexandre Sablayrolles}, \bibinfo{person}{Antoine Roux}, \bibinfo{person}{Arthur Mensch}, \bibinfo{person}{Blanche Savary}, \bibinfo{person}{Chris Bamford}, \bibinfo{person}{Devendra~Singh Chaplot}, \bibinfo{person}{Diego de~las Casas}, \bibinfo{person}{Emma~Bou Hanna}, \bibinfo{person}{Florian Bressand}, {et~al\mbox{.}}} \bibinfo{year}{2024}\natexlab{}.
\newblock \showarticletitle{Mixtral of experts}.
\newblock \bibinfo{journal}{\emph{arXiv preprint arXiv:2401.04088}} (\bibinfo{year}{2024}).
\newblock


\bibitem[Kwon et~al\mbox{.}(2023)]%
        {kwon2023efficient}
\bibfield{author}{\bibinfo{person}{Woosuk Kwon}, \bibinfo{person}{Zhuohan Li}, \bibinfo{person}{Siyuan Zhuang}, \bibinfo{person}{Ying Sheng}, \bibinfo{person}{Lianmin Zheng}, \bibinfo{person}{Cody~Hao Yu}, \bibinfo{person}{Joseph Gonzalez}, \bibinfo{person}{Hao Zhang}, {and} \bibinfo{person}{Ion Stoica}.} \bibinfo{year}{2023}\natexlab{}.
\newblock \showarticletitle{Efficient memory management for large language model serving with pagedattention}. In \bibinfo{booktitle}{\emph{Proceedings of the 29th Symposium on Operating Systems Principles}}. \bibinfo{pages}{611--626}.
\newblock


\bibitem[Le~Scao et~al\mbox{.}(2023)]%
        {le2023bloom}
\bibfield{author}{\bibinfo{person}{Teven Le~Scao}, \bibinfo{person}{Angela Fan}, \bibinfo{person}{Christopher Akiki}, \bibinfo{person}{Ellie Pavlick}, \bibinfo{person}{Suzana Ili{\'c}}, \bibinfo{person}{Daniel Hesslow}, \bibinfo{person}{Roman Castagn{\'e}}, \bibinfo{person}{Alexandra~Sasha Luccioni}, \bibinfo{person}{Fran{\c{c}}ois Yvon}, \bibinfo{person}{Matthias Gall{\'e}}, {et~al\mbox{.}}} \bibinfo{year}{2023}\natexlab{}.
\newblock \showarticletitle{Bloom: A 176b-parameter open-access multilingual language model}.
\newblock  (\bibinfo{year}{2023}).
\newblock


\bibitem[Lee et~al\mbox{.}(2023)]%
        {lee2023rlaif}
\bibfield{author}{\bibinfo{person}{Harrison Lee}, \bibinfo{person}{Samrat Phatale}, \bibinfo{person}{Hassan Mansoor}, \bibinfo{person}{Kellie Lu}, \bibinfo{person}{Thomas Mesnard}, \bibinfo{person}{Colton Bishop}, \bibinfo{person}{Victor Carbune}, {and} \bibinfo{person}{Abhinav Rastogi}.} \bibinfo{year}{2023}\natexlab{}.
\newblock \showarticletitle{Rlaif: Scaling reinforcement learning from human feedback with ai feedback}.
\newblock \bibinfo{journal}{\emph{arXiv preprint arXiv:2309.00267}} (\bibinfo{year}{2023}).
\newblock


\bibitem[Levshun(2023)]%
        {levshun2023comparative}
\bibfield{author}{\bibinfo{person}{Dmitry Levshun}.} \bibinfo{year}{2023}\natexlab{}.
\newblock \showarticletitle{Comparative Analysis of Machine Learning Methods in Vulnerability Metrics Transformation}. In \bibinfo{booktitle}{\emph{International Conference on Intelligent Information Technologies for Industry}}. Springer, \bibinfo{pages}{60--70}.
\newblock


\bibitem[Lewis et~al\mbox{.}(2020)]%
        {lewis2020retrieval}
\bibfield{author}{\bibinfo{person}{Patrick Lewis}, \bibinfo{person}{Ethan Perez}, \bibinfo{person}{Aleksandra Piktus}, \bibinfo{person}{Fabio Petroni}, \bibinfo{person}{Vladimir Karpukhin}, \bibinfo{person}{Naman Goyal}, \bibinfo{person}{Heinrich K{\"u}ttler}, \bibinfo{person}{Mike Lewis}, \bibinfo{person}{Wen-tau Yih}, \bibinfo{person}{Tim Rockt{\"a}schel}, {et~al\mbox{.}}} \bibinfo{year}{2020}\natexlab{}.
\newblock \showarticletitle{Retrieval-augmented generation for knowledge-intensive nlp tasks}.
\newblock \bibinfo{journal}{\emph{Advances in Neural Information Processing Systems}}  \bibinfo{volume}{33} (\bibinfo{year}{2020}), \bibinfo{pages}{9459--9474}.
\newblock


\bibitem[Li et~al\mbox{.}(2023)]%
        {li2023large}
\bibfield{author}{\bibinfo{person}{Yinheng Li}, \bibinfo{person}{Shaofei Wang}, \bibinfo{person}{Han Ding}, {and} \bibinfo{person}{Hang Chen}.} \bibinfo{year}{2023}\natexlab{}.
\newblock \showarticletitle{Large language models in finance: A survey}. In \bibinfo{booktitle}{\emph{Proceedings of the Fourth ACM International Conference on AI in Finance}}. \bibinfo{pages}{374--382}.
\newblock


\bibitem[Lin and Och(2004)]%
        {lin2004looking}
\bibfield{author}{\bibinfo{person}{Chin-Yew Lin} {and} \bibinfo{person}{FJ Och}.} \bibinfo{year}{2004}\natexlab{}.
\newblock \showarticletitle{Looking for a few good metrics: ROUGE and its evaluation}. In \bibinfo{booktitle}{\emph{Ntcir workshop}}.
\newblock


\bibitem[Liu et~al\mbox{.}(2023)]%
        {liu2023not}
\bibfield{author}{\bibinfo{person}{Xin Liu}, \bibinfo{person}{Yuan Tan}, \bibinfo{person}{Zhenghang Xiao}, \bibinfo{person}{Jianwei Zhuge}, {and} \bibinfo{person}{Rui Zhou}.} \bibinfo{year}{2023}\natexlab{}.
\newblock \showarticletitle{Not the end of story: An evaluation of chatgpt-driven vulnerability description mappings}. In \bibinfo{booktitle}{\emph{Findings of the Association for Computational Linguistics: ACL 2023}}. \bibinfo{pages}{3724--3731}.
\newblock


\bibitem[Liu et~al\mbox{.}(2019)]%
        {liu2019roberta}
\bibfield{author}{\bibinfo{person}{Yinhan Liu}, \bibinfo{person}{Myle Ott}, \bibinfo{person}{Naman Goyal}, \bibinfo{person}{Jingfei Du}, \bibinfo{person}{Mandar Joshi}, \bibinfo{person}{Danqi Chen}, \bibinfo{person}{Omer Levy}, \bibinfo{person}{Mike Lewis}, \bibinfo{person}{Luke Zettlemoyer}, {and} \bibinfo{person}{Veselin Stoyanov}.} \bibinfo{year}{2019}\natexlab{}.
\newblock \showarticletitle{Roberta: A robustly optimized bert pretraining approach}.
\newblock \bibinfo{journal}{\emph{arXiv preprint arXiv:1907.11692}} (\bibinfo{year}{2019}).
\newblock


\bibitem[Lu et~al\mbox{.}(2024)]%
        {lu2024grace}
\bibfield{author}{\bibinfo{person}{Guilong Lu}, \bibinfo{person}{Xiaolin Ju}, \bibinfo{person}{Xiang Chen}, \bibinfo{person}{Wenlong Pei}, {and} \bibinfo{person}{Zhilong Cai}.} \bibinfo{year}{2024}\natexlab{}.
\newblock \showarticletitle{GRACE: Empowering LLM-based software vulnerability detection with graph structure and in-context learning}.
\newblock \bibinfo{journal}{\emph{Journal of Systems and Software}}  \bibinfo{volume}{212} (\bibinfo{year}{2024}), \bibinfo{pages}{112031}.
\newblock


\bibitem[Lu et~al\mbox{.}(2023)]%
        {lu2023routing}
\bibfield{author}{\bibinfo{person}{Keming Lu}, \bibinfo{person}{Hongyi Yuan}, \bibinfo{person}{Runji Lin}, \bibinfo{person}{Junyang Lin}, \bibinfo{person}{Zheng Yuan}, \bibinfo{person}{Chang Zhou}, {and} \bibinfo{person}{Jingren Zhou}.} \bibinfo{year}{2023}\natexlab{}.
\newblock \showarticletitle{Routing to the expert: Efficient reward-guided ensemble of large language models}.
\newblock \bibinfo{journal}{\emph{arXiv preprint arXiv:2311.08692}} (\bibinfo{year}{2023}).
\newblock


\bibitem[Mell et~al\mbox{.}(2006)]%
        {mell2006common}
\bibfield{author}{\bibinfo{person}{Peter Mell}, \bibinfo{person}{Karen Scarfone}, {and} \bibinfo{person}{Sasha Romanosky}.} \bibinfo{year}{2006}\natexlab{}.
\newblock \showarticletitle{Common vulnerability scoring system}.
\newblock \bibinfo{journal}{\emph{IEEE Security \& Privacy}} \bibinfo{volume}{4}, \bibinfo{number}{6} (\bibinfo{year}{2006}), \bibinfo{pages}{85--89}.
\newblock


\bibitem[Mitre(2005)]%
        {mitre2005common}
\bibfield{author}{\bibinfo{person}{CVE Mitre}.} \bibinfo{year}{2005}\natexlab{}.
\newblock \bibinfo{title}{Common vulnerabilities and exposures}.
\newblock
\newblock


\bibitem[Moore et~al\mbox{.}(2023)]%
        {moore2023empowering}
\bibfield{author}{\bibinfo{person}{Steven Moore}, \bibinfo{person}{Richard Tong}, \bibinfo{person}{Anjali Singh}, \bibinfo{person}{Zitao Liu}, \bibinfo{person}{Xiangen Hu}, \bibinfo{person}{Yu Lu}, \bibinfo{person}{Joleen Liang}, \bibinfo{person}{Chen Cao}, \bibinfo{person}{Hassan Khosravi}, \bibinfo{person}{Paul Denny}, {et~al\mbox{.}}} \bibinfo{year}{2023}\natexlab{}.
\newblock \showarticletitle{Empowering education with llms-the next-gen interface and content generation}. In \bibinfo{booktitle}{\emph{International Conference on Artificial Intelligence in Education}}. Springer, \bibinfo{pages}{32--37}.
\newblock


\bibitem[Naito et~al\mbox{.}(2023a)]%
        {naito2023llm}
\bibfield{author}{\bibinfo{person}{Takeru Naito}, \bibinfo{person}{Rei Watanabe}, {and} \bibinfo{person}{Takuho Mitsunaga}.} \bibinfo{year}{2023}\natexlab{a}.
\newblock \showarticletitle{LLM-based Attack Scenarios Generator with IT Asset Management and Vulnerability Information}. In \bibinfo{booktitle}{\emph{2023 6th International Conference on Signal Processing and Information Security (ICSPIS)}}. IEEE, \bibinfo{pages}{99--103}.
\newblock


\bibitem[Naito et~al\mbox{.}(2023b)]%
        {llm_it}
\bibfield{author}{\bibinfo{person}{Takeru Naito}, \bibinfo{person}{Rei Watanabe}, {and} \bibinfo{person}{Takuho Mitsunaga}.} \bibinfo{year}{2023}\natexlab{b}.
\newblock \showarticletitle{LLM-based Attack Scenarios Generator with IT Asset Management and Vulnerability Information}. In \bibinfo{booktitle}{\emph{2023 6th International Conference on Signal Processing and Information Security (ICSPIS)}}. \bibinfo{pages}{99--103}.
\newblock
\urldef\tempurl%
\url{https://doi.org/10.1109/ICSPIS60075.2023.10344019}
\showDOI{\tempurl}


\bibitem[Naveed et~al\mbox{.}(2023)]%
        {naveed2023comprehensive}
\bibfield{author}{\bibinfo{person}{Humza Naveed}, \bibinfo{person}{Asad~Ullah Khan}, \bibinfo{person}{Shi Qiu}, \bibinfo{person}{Muhammad Saqib}, \bibinfo{person}{Saeed Anwar}, \bibinfo{person}{Muhammad Usman}, \bibinfo{person}{Nick Barnes}, {and} \bibinfo{person}{Ajmal Mian}.} \bibinfo{year}{2023}\natexlab{}.
\newblock \showarticletitle{A comprehensive overview of large language models}.
\newblock \bibinfo{journal}{\emph{arXiv preprint arXiv:2307.06435}} (\bibinfo{year}{2023}).
\newblock


\bibitem[of~Health et~al\mbox{.}(2016)]%
        {fdareport}
\bibfield{author}{\bibinfo{person}{U.S.~Department of Health}, \bibinfo{person}{Food Human~Services}, \bibinfo{person}{Center for~Devices Drug~Administration}, \bibinfo{person}{Center for Biologics~Evaluation Radiological~Health, Office of the Center~Director}, {and} \bibinfo{person}{Research}.} \bibinfo{year}{December 28, 2016}\natexlab{}.
\newblock \bibinfo{booktitle}{\emph{Postmarket Management of Cybersecurity in Medical Devices: Guidance for Industry and Food and Drug Administration Staff}}.
\newblock
\urldef\tempurl%
\url{https://www.fda.gov/media/95862/download}
\showURL{%
\tempurl}


\bibitem[OpenAI(2021)]%
        {chatgpt}
\bibfield{author}{\bibinfo{person}{OpenAI}.} \bibinfo{year}{2021}\natexlab{}.
\newblock \bibinfo{title}{ChatGPT}.
\newblock \bibinfo{howpublished}{\url{https://openai.com/gpt-3/}}.
\newblock


\bibitem[Pearce et~al\mbox{.}(2023)]%
        {pearce2023examining}
\bibfield{author}{\bibinfo{person}{Hammond Pearce}, \bibinfo{person}{Benjamin Tan}, \bibinfo{person}{Baleegh Ahmad}, \bibinfo{person}{Ramesh Karri}, {and} \bibinfo{person}{Brendan Dolan-Gavitt}.} \bibinfo{year}{2023}\natexlab{}.
\newblock \showarticletitle{Examining zero-shot vulnerability repair with large language models}. In \bibinfo{booktitle}{\emph{2023 IEEE Symposium on Security and Privacy (SP)}}. IEEE, \bibinfo{pages}{2339--2356}.
\newblock


\bibitem[Pedregosa et~al\mbox{.}(2011)]%
        {scikit-learn}
\bibfield{author}{\bibinfo{person}{F. Pedregosa}, \bibinfo{person}{G. Varoquaux}, \bibinfo{person}{A. Gramfort}, \bibinfo{person}{V. Michel}, \bibinfo{person}{B. Thirion}, \bibinfo{person}{O. Grisel}, \bibinfo{person}{M. Blondel}, \bibinfo{person}{P. Prettenhofer}, \bibinfo{person}{R. Weiss}, \bibinfo{person}{V. Dubourg}, \bibinfo{person}{J. Vanderplas}, \bibinfo{person}{A. Passos}, \bibinfo{person}{D. Cournapeau}, \bibinfo{person}{M. Brucher}, \bibinfo{person}{M. Perrot}, {and} \bibinfo{person}{E. Duchesnay}.} \bibinfo{year}{2011}\natexlab{}.
\newblock \showarticletitle{Scikit-learn: Machine Learning in {P}ython}.
\newblock \bibinfo{journal}{\emph{Journal of Machine Learning Research}}  \bibinfo{volume}{12} (\bibinfo{year}{2011}), \bibinfo{pages}{2825--2830}.
\newblock


\bibitem[Prasad et~al\mbox{.}(2018)]%
        {prasad2018integrated}
\bibfield{author}{\bibinfo{person}{K~Siva Prasad}, \bibinfo{person}{K~Raja Sekhar}, {and} \bibinfo{person}{P Rajarajeswari}.} \bibinfo{year}{2018}\natexlab{}.
\newblock \showarticletitle{An integrated approach towards vulnerability assessment \& penetration testing for a web application}.
\newblock \bibinfo{journal}{\emph{International Journal of Engineering and Technology (UAE)}}  \bibinfo{volume}{7} (\bibinfo{year}{2018}), \bibinfo{pages}{431--435}.
\newblock


\bibitem[Press et~al\mbox{.}(2021)]%
        {press2021train}
\bibfield{author}{\bibinfo{person}{Ofir Press}, \bibinfo{person}{Noah~A Smith}, {and} \bibinfo{person}{Mike Lewis}.} \bibinfo{year}{2021}\natexlab{}.
\newblock \showarticletitle{Train short, test long: Attention with linear biases enables input length extrapolation}.
\newblock \bibinfo{journal}{\emph{arXiv preprint arXiv:2108.12409}} (\bibinfo{year}{2021}).
\newblock


\bibitem[Radford et~al\mbox{.}(2018)]%
        {radford2018improving}
\bibfield{author}{\bibinfo{person}{Alec Radford}, \bibinfo{person}{Karthik Narasimhan}, \bibinfo{person}{Tim Salimans}, \bibinfo{person}{Ilya Sutskever}, {et~al\mbox{.}}} \bibinfo{year}{2018}\natexlab{}.
\newblock \showarticletitle{Improving language understanding by generative pre-training}.
\newblock  (\bibinfo{year}{2018}).
\newblock


\bibitem[Rasley et~al\mbox{.}(2020)]%
        {rasley2020deepspeed}
\bibfield{author}{\bibinfo{person}{Jeff Rasley}, \bibinfo{person}{Samyam Rajbhandari}, \bibinfo{person}{Olatunji Ruwase}, {and} \bibinfo{person}{Yuxiong He}.} \bibinfo{year}{2020}\natexlab{}.
\newblock \showarticletitle{Deepspeed: System optimizations enable training deep learning models with over 100 billion parameters}. In \bibinfo{booktitle}{\emph{Proceedings of the 26th ACM SIGKDD International Conference on Knowledge Discovery \& Data Mining}}. \bibinfo{pages}{3505--3506}.
\newblock


\bibitem[Sallam(2023)]%
        {sallam2023chatgpt}
\bibfield{author}{\bibinfo{person}{Malik Sallam}.} \bibinfo{year}{2023}\natexlab{}.
\newblock \showarticletitle{ChatGPT utility in healthcare education, research, and practice: systematic review on the promising perspectives and valid concerns}. In \bibinfo{booktitle}{\emph{Healthcare}}, Vol.~\bibinfo{volume}{11}. MDPI, \bibinfo{pages}{887}.
\newblock


\bibitem[Sewak et~al\mbox{.}(2023)]%
        {sewak2023crush}
\bibfield{author}{\bibinfo{person}{Mohit Sewak}, \bibinfo{person}{Vamsi Emani}, {and} \bibinfo{person}{Annam Naresh}.} \bibinfo{year}{2023}\natexlab{}.
\newblock \showarticletitle{CRUSH: Cybersecurity Research using Universal LLMs and Semantic Hypernetworks}.
\newblock  (\bibinfo{year}{2023}).
\newblock


\bibitem[Shahid and Debar(2021)]%
        {shahid2021cvss}
\bibfield{author}{\bibinfo{person}{Mustafizur~R Shahid} {and} \bibinfo{person}{Herv{\'e} Debar}.} \bibinfo{year}{2021}\natexlab{}.
\newblock \showarticletitle{Cvss-bert: Explainable natural language processing to determine the severity of a computer security vulnerability from its description}. In \bibinfo{booktitle}{\emph{2021 20th IEEE International Conference on Machine Learning and Applications (ICMLA)}}. IEEE, \bibinfo{pages}{1600--1607}.
\newblock


\bibitem[Steenhoek et~al\mbox{.}(2024)]%
        {steenhoek2024comprehensive}
\bibfield{author}{\bibinfo{person}{Benjamin Steenhoek}, \bibinfo{person}{Md~Mahbubur Rahman}, \bibinfo{person}{Monoshi~Kumar Roy}, \bibinfo{person}{Mirza~Sanjida Alam}, \bibinfo{person}{Earl~T Barr}, {and} \bibinfo{person}{Wei Le}.} \bibinfo{year}{2024}\natexlab{}.
\newblock \showarticletitle{A Comprehensive Study of the Capabilities of Large Language Models for Vulnerability Detection}.
\newblock \bibinfo{journal}{\emph{arXiv preprint arXiv:2403.17218}} (\bibinfo{year}{2024}).
\newblock


\bibitem[Sun et~al\mbox{.}(2023)]%
        {sun2023text}
\bibfield{author}{\bibinfo{person}{Xiaofei Sun}, \bibinfo{person}{Xiaoya Li}, \bibinfo{person}{Jiwei Li}, \bibinfo{person}{Fei Wu}, \bibinfo{person}{Shangwei Guo}, \bibinfo{person}{Tianwei Zhang}, {and} \bibinfo{person}{Guoyin Wang}.} \bibinfo{year}{2023}\natexlab{}.
\newblock \showarticletitle{Text classification via large language models}.
\newblock \bibinfo{journal}{\emph{arXiv preprint arXiv:2305.08377}} (\bibinfo{year}{2023}).
\newblock


\bibitem[Sun et~al\mbox{.}(2024)]%
        {sun2024llm4vuln}
\bibfield{author}{\bibinfo{person}{Yuqiang Sun}, \bibinfo{person}{Daoyuan Wu}, \bibinfo{person}{Yue Xue}, \bibinfo{person}{Han Liu}, \bibinfo{person}{Wei Ma}, \bibinfo{person}{Lyuye Zhang}, \bibinfo{person}{Miaolei Shi}, {and} \bibinfo{person}{Yang Liu}.} \bibinfo{year}{2024}\natexlab{}.
\newblock \showarticletitle{LLM4Vuln: A Unified Evaluation Framework for Decoupling and Enhancing LLMs' Vulnerability Reasoning}.
\newblock \bibinfo{journal}{\emph{arXiv preprint arXiv:2401.16185}} (\bibinfo{year}{2024}).
\newblock


\bibitem[Tan et~al\mbox{.}(2024)]%
        {tan2024llm4decompile}
\bibfield{author}{\bibinfo{person}{Hanzhuo Tan}, \bibinfo{person}{Qi Luo}, \bibinfo{person}{Jing Li}, {and} \bibinfo{person}{Yuqun Zhang}.} \bibinfo{year}{2024}\natexlab{}.
\newblock \showarticletitle{LLM4Decompile: Decompiling Binary Code with Large Language Models}.
\newblock \bibinfo{journal}{\emph{arXiv preprint arXiv:2403.05286}} (\bibinfo{year}{2024}).
\newblock


\bibitem[Taori et~al\mbox{.}(2023)]%
        {taori2023alpaca}
\bibfield{author}{\bibinfo{person}{Rohan Taori}, \bibinfo{person}{Ishaan Gulrajani}, \bibinfo{person}{Tianyi Zhang}, \bibinfo{person}{Yann Dubois}, \bibinfo{person}{Xuechen Li}, \bibinfo{person}{Carlos Guestrin}, \bibinfo{person}{Percy Liang}, {and} \bibinfo{person}{Tatsunori~B Hashimoto}.} \bibinfo{year}{2023}\natexlab{}.
\newblock \showarticletitle{Alpaca: A strong, replicable instruction-following model}.
\newblock \bibinfo{journal}{\emph{Stanford Center for Research on Foundation Models. https://crfm. stanford. edu/2023/03/13/alpaca. html}} \bibinfo{volume}{3}, \bibinfo{number}{6} (\bibinfo{year}{2023}), \bibinfo{pages}{7}.
\newblock


\bibitem[Team(2023)]%
        {MosaicML2023Introducing}
\bibfield{author}{\bibinfo{person}{MosaicML~NLP Team}.} \bibinfo{year}{2023}\natexlab{}.
\newblock \bibinfo{booktitle}{\emph{Introducing MPT-7B: A New Standard for Open-Source, Commercially Usable LLMs}}.
\newblock
\urldef\tempurl%
\url{www.mosaicml.com/blog/mpt-7b}
\showURL{%
\tempurl}
\newblock
\shownote{Accessed: 2023-05-05}.


\bibitem[Touvron et~al\mbox{.}(2023)]%
        {touvron2023llama}
\bibfield{author}{\bibinfo{person}{Hugo Touvron}, \bibinfo{person}{Thibaut Lavril}, \bibinfo{person}{Gautier Izacard}, \bibinfo{person}{Xavier Martinet}, \bibinfo{person}{Marie-Anne Lachaux}, \bibinfo{person}{Timoth{\'e}e Lacroix}, \bibinfo{person}{Baptiste Rozi{\`e}re}, \bibinfo{person}{Naman Goyal}, \bibinfo{person}{Eric Hambro}, \bibinfo{person}{Faisal Azhar}, {et~al\mbox{.}}} \bibinfo{year}{2023}\natexlab{}.
\newblock \showarticletitle{Llama: Open and efficient foundation language models}.
\newblock \bibinfo{journal}{\emph{arXiv preprint arXiv:2302.13971}} (\bibinfo{year}{2023}).
\newblock


\bibitem[Vasireddy et~al\mbox{.}(2023)]%
        {vasireddy2023cvss}
\bibfield{author}{\bibinfo{person}{Dinesh~T Vasireddy}, \bibinfo{person}{Dakota~S Dale}, {and} \bibinfo{person}{Qinghua Li}.} \bibinfo{year}{2023}\natexlab{}.
\newblock \showarticletitle{CVSS Base Score Prediction Using an Optimized Machine Learning Scheme}. In \bibinfo{booktitle}{\emph{2023 Resilience Week (RWS)}}. IEEE, \bibinfo{pages}{1--6}.
\newblock


\bibitem[von Werra et~al\mbox{.}(2020)]%
        {vonwerra2022trl}
\bibfield{author}{\bibinfo{person}{Leandro von Werra}, \bibinfo{person}{Younes Belkada}, \bibinfo{person}{Lewis Tunstall}, \bibinfo{person}{Edward Beeching}, \bibinfo{person}{Tristan Thrush}, \bibinfo{person}{Nathan Lambert}, {and} \bibinfo{person}{Shengyi Huang}.} \bibinfo{year}{2020}\natexlab{}.
\newblock \bibinfo{title}{TRL: Transformer Reinforcement Learning}.
\newblock \bibinfo{howpublished}{\url{https://github.com/huggingface/trl}}.
\newblock


\bibitem[Wei et~al\mbox{.}(2022)]%
        {wei2022chain}
\bibfield{author}{\bibinfo{person}{Jason Wei}, \bibinfo{person}{Xuezhi Wang}, \bibinfo{person}{Dale Schuurmans}, \bibinfo{person}{Maarten Bosma}, \bibinfo{person}{Fei Xia}, \bibinfo{person}{Ed Chi}, \bibinfo{person}{Quoc~V Le}, \bibinfo{person}{Denny Zhou}, {et~al\mbox{.}}} \bibinfo{year}{2022}\natexlab{}.
\newblock \showarticletitle{Chain-of-thought prompting elicits reasoning in large language models}.
\newblock \bibinfo{journal}{\emph{Advances in neural information processing systems}}  \bibinfo{volume}{35} (\bibinfo{year}{2022}), \bibinfo{pages}{24824--24837}.
\newblock


\bibitem[Whitehouse et~al\mbox{.}(2023)]%
        {whitehouse2023llm}
\bibfield{author}{\bibinfo{person}{Chenxi Whitehouse}, \bibinfo{person}{Monojit Choudhury}, {and} \bibinfo{person}{Alham~Fikri Aji}.} \bibinfo{year}{2023}\natexlab{}.
\newblock \showarticletitle{Llm-powered data augmentation for enhanced crosslingual performance}.
\newblock \bibinfo{journal}{\emph{arXiv preprint arXiv:2305.14288}} (\bibinfo{year}{2023}).
\newblock


\bibitem[Wu et~al\mbox{.}(2023)]%
        {wu2023autogen}
\bibfield{author}{\bibinfo{person}{Qingyun Wu}, \bibinfo{person}{Gagan Bansal}, \bibinfo{person}{Jieyu Zhang}, \bibinfo{person}{Yiran Wu}, \bibinfo{person}{Shaokun Zhang}, \bibinfo{person}{Erkang Zhu}, \bibinfo{person}{Beibin Li}, \bibinfo{person}{Li Jiang}, \bibinfo{person}{Xiaoyun Zhang}, {and} \bibinfo{person}{Chi Wang}.} \bibinfo{year}{2023}\natexlab{}.
\newblock \showarticletitle{Autogen: Enabling next-gen llm applications via multi-agent conversation framework}.
\newblock \bibinfo{journal}{\emph{arXiv preprint arXiv:2308.08155}} (\bibinfo{year}{2023}).
\newblock


\bibitem[Yang et~al\mbox{.}(2024)]%
        {yang2024large}
\bibfield{author}{\bibinfo{person}{Xiaoxian Yang}, \bibinfo{person}{Zhifeng Wang}, \bibinfo{person}{Qi Wang}, \bibinfo{person}{Ke Wei}, \bibinfo{person}{Kaiqi Zhang}, {and} \bibinfo{person}{Jiangang Shi}.} \bibinfo{year}{2024}\natexlab{}.
\newblock \showarticletitle{Large language models for automated q\&a involving legal documents: a survey on algorithms, frameworks and applications}.
\newblock \bibinfo{journal}{\emph{International Journal of Web Information Systems}} (\bibinfo{year}{2024}).
\newblock


\bibitem[Yao et~al\mbox{.}(2024)]%
        {yao2024survey}
\bibfield{author}{\bibinfo{person}{Yifan Yao}, \bibinfo{person}{Jinhao Duan}, \bibinfo{person}{Kaidi Xu}, \bibinfo{person}{Yuanfang Cai}, \bibinfo{person}{Zhibo Sun}, {and} \bibinfo{person}{Yue Zhang}.} \bibinfo{year}{2024}\natexlab{}.
\newblock \showarticletitle{A survey on large language model (llm) security and privacy: The good, the bad, and the ugly}.
\newblock \bibinfo{journal}{\emph{High-Confidence Computing}} (\bibinfo{year}{2024}), \bibinfo{pages}{100211}.
\newblock


\bibitem[Yin et~al\mbox{.}(2020)]%
        {yin2020apply}
\bibfield{author}{\bibinfo{person}{Jiao Yin}, \bibinfo{person}{MingJian Tang}, \bibinfo{person}{Jinli Cao}, {and} \bibinfo{person}{Hua Wang}.} \bibinfo{year}{2020}\natexlab{}.
\newblock \showarticletitle{Apply transfer learning to cybersecurity: Predicting exploitability of vulnerabilities by description}.
\newblock \bibinfo{journal}{\emph{Knowledge-Based Systems}}  \bibinfo{volume}{210} (\bibinfo{year}{2020}), \bibinfo{pages}{106529}.
\newblock


\bibitem[Yin and Ni(2024)]%
        {yin2024multitask}
\bibfield{author}{\bibinfo{person}{Xin Yin} {and} \bibinfo{person}{Chao Ni}.} \bibinfo{year}{2024}\natexlab{}.
\newblock \showarticletitle{Multitask-based Evaluation of Open-Source LLM on Software Vulnerability}.
\newblock \bibinfo{journal}{\emph{arXiv preprint arXiv:2404.02056}} (\bibinfo{year}{2024}).
\newblock


\bibitem[Yosifova et~al\mbox{.}(2021)]%
        {yosifova2021predicting}
\bibfield{author}{\bibinfo{person}{Veneta Yosifova}, \bibinfo{person}{Antoniya Tasheva}, {and} \bibinfo{person}{Roumen Trifonov}.} \bibinfo{year}{2021}\natexlab{}.
\newblock \showarticletitle{Predicting vulnerability type in common vulnerabilities and exposures (cve) database with machine learning classifiers}. In \bibinfo{booktitle}{\emph{2021 12th National Conference with International Participation (ELECTRONICA)}}. IEEE, \bibinfo{pages}{1--6}.
\newblock


\bibitem[Zhang et~al\mbox{.}(2024b)]%
        {zhang2024vtt}
\bibfield{author}{\bibinfo{person}{Chenhui Zhang}, \bibinfo{person}{Le Wang}, \bibinfo{person}{Dunqiu Fan}, \bibinfo{person}{Junyi Zhu}, \bibinfo{person}{Tang Zhou}, \bibinfo{person}{Liyi Zeng}, {and} \bibinfo{person}{Zhaohua Li}.} \bibinfo{year}{2024}\natexlab{b}.
\newblock \showarticletitle{VTT-LLM: Advancing Vulnerability-to-Tactic-and-Technique Mapping through Fine-Tuning of Large Language Model}.
\newblock \bibinfo{journal}{\emph{Mathematics}} \bibinfo{volume}{12}, \bibinfo{number}{9} (\bibinfo{year}{2024}), \bibinfo{pages}{1286}.
\newblock


\bibitem[Zhang et~al\mbox{.}(2024a)]%
        {zhang2024llms}
\bibfield{author}{\bibinfo{person}{Jie Zhang}, \bibinfo{person}{Haoyu Bu}, \bibinfo{person}{Hui Wen}, \bibinfo{person}{Yu Chen}, \bibinfo{person}{Lun Li}, {and} \bibinfo{person}{Hongsong Zhu}.} \bibinfo{year}{2024}\natexlab{a}.
\newblock \showarticletitle{When LLMs Meet Cybersecurity: A Systematic Literature Review}.
\newblock \bibinfo{journal}{\emph{arXiv preprint arXiv:2405.03644}} (\bibinfo{year}{2024}).
\newblock


\bibitem[Zhou et~al\mbox{.}(2024)]%
        {zhou2024large}
\bibfield{author}{\bibinfo{person}{Xin Zhou}, \bibinfo{person}{Sicong Cao}, \bibinfo{person}{Xiaobing Sun}, {and} \bibinfo{person}{David Lo}.} \bibinfo{year}{2024}\natexlab{}.
\newblock \showarticletitle{Large Language Model for Vulnerability Detection and Repair: Literature Review and Roadmap}.
\newblock \bibinfo{journal}{\emph{arXiv preprint arXiv:2404.02525}} (\bibinfo{year}{2024}).
\newblock


\end{thebibliography}

\appendix

\begin{table*}
      \begin{tabular}{p{0.15\linewidth} | p{0.4\linewidth} | p{0.4\linewidth}}
        \toprule
        Type of Error&Expected Response&Generated Response\\
        \midrule
        Wrong generation & The issue may be addressed in a future security patch or software version. &Vulnerability is only exploitable in context of ``prohibited” use cases\\
        \hline
        Omission & CoreMedia, WebRTC is not utilized in the product.&Technology component is not deployed. \\
        \hline
        Wrong key information & .NET 7.0 is not deployed and utilized in the product.& .NET 6.0 is not deployed and utilized in the product. \\
      \bottomrule
    \end{tabular}
    \caption{Common errors in CVE-LLM generation}
      \label{tab:generr}
    \end{table*}

\section{Environmental Vector test dataset results}

\begin{figure*}
    \centering
    \begin{subfigure}[b]{0.3\textwidth}
        \includegraphics[width=\textwidth]{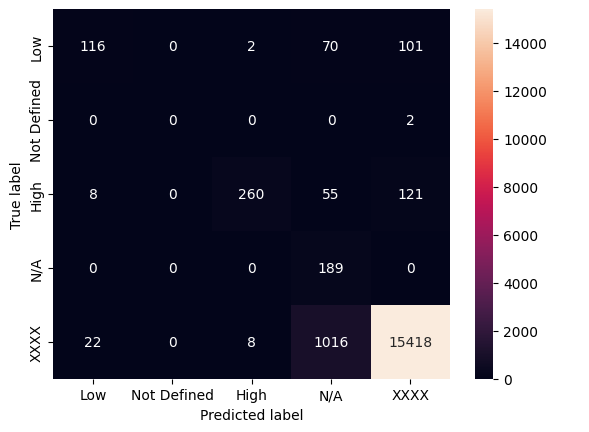}
        \caption{Modified Attack Complexity}
    \end{subfigure}
    ~ 
    \begin{subfigure}[b]{0.3\textwidth}
        \includegraphics[width=\textwidth]{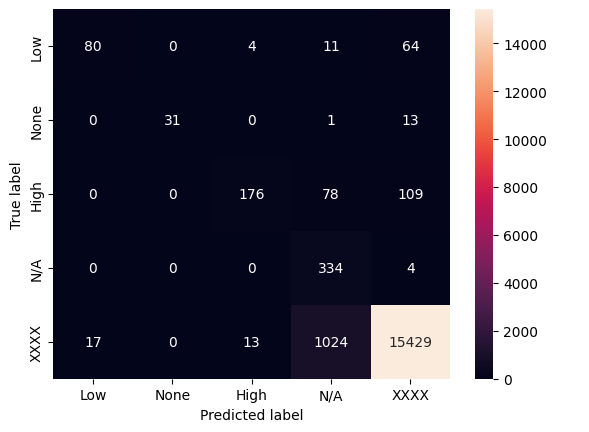}
        \caption{Modified Availability}
    \end{subfigure}
    ~ 
    \begin{subfigure}[b]{0.3\textwidth}
        \includegraphics[width=\textwidth]{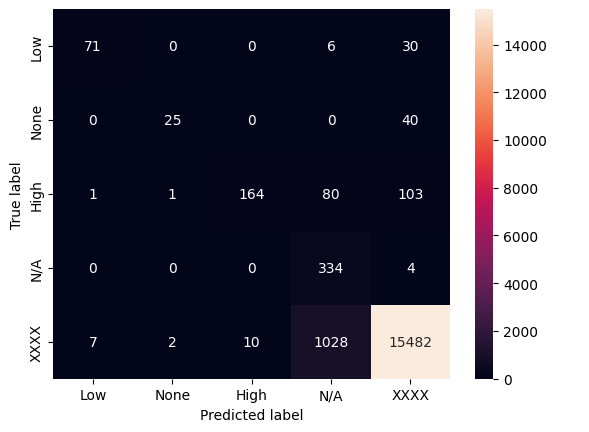}
        \caption{Modified Confidentiality}
    \end{subfigure}
    \hfill
     \begin{subfigure}[b]{0.3\textwidth}
        \includegraphics[width=\textwidth]{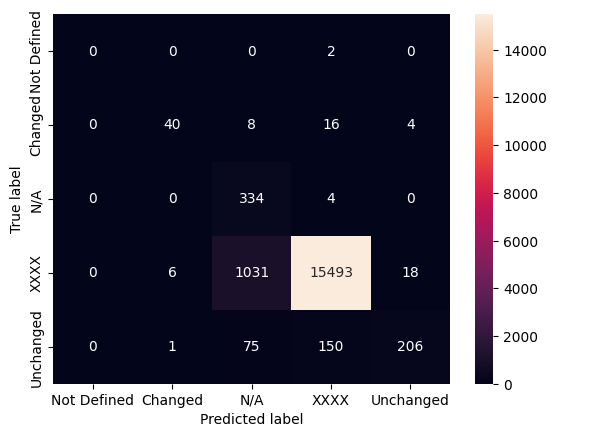}
        \caption{Modified Scope}
    \end{subfigure}
    ~ 
    \begin{subfigure}[b]{0.3\textwidth}
        \includegraphics[width=\textwidth]{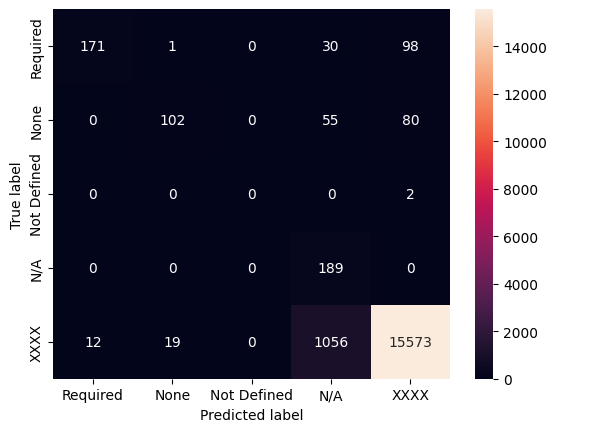}
        \caption{Modified User Interaction}
    \end{subfigure}
    ~ 
    \begin{subfigure}[b]{0.3\textwidth}
        \includegraphics[width=\textwidth]{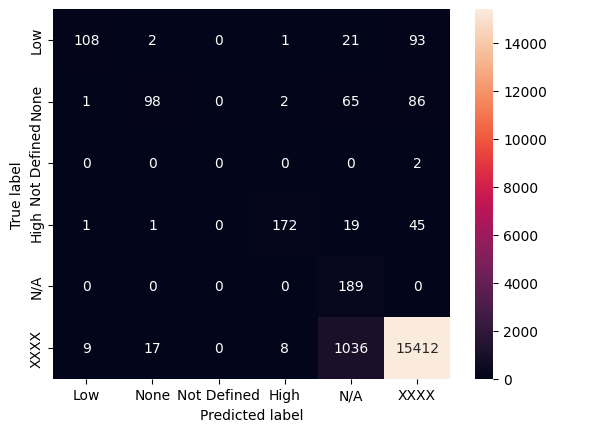}
        \caption{Modified Privileges Required}
    \end{subfigure}
    \hfill
    \begin{subfigure}[b]{0.3\textwidth}
        \includegraphics[width=\textwidth]{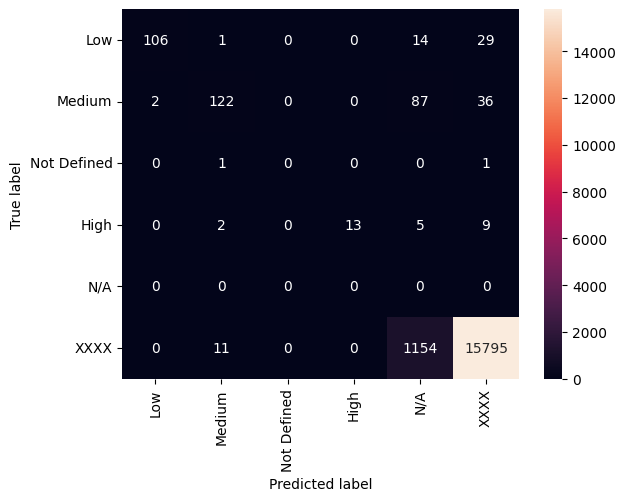}
        \caption{Confidentiality Requirement}
    \end{subfigure}
    ~ 
    \begin{subfigure}[b]{0.3\textwidth}
        \includegraphics[width=\textwidth]{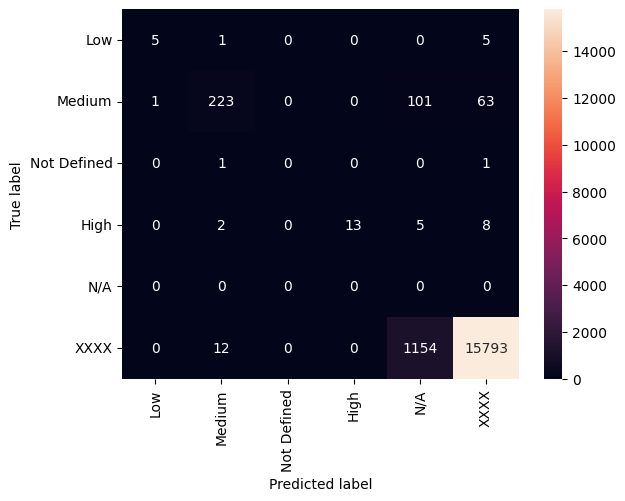}
        \caption{Integrity Requirement}
    \end{subfigure}
    ~ 
    \begin{subfigure}[b]{0.3\textwidth}
        \includegraphics[width=\textwidth]{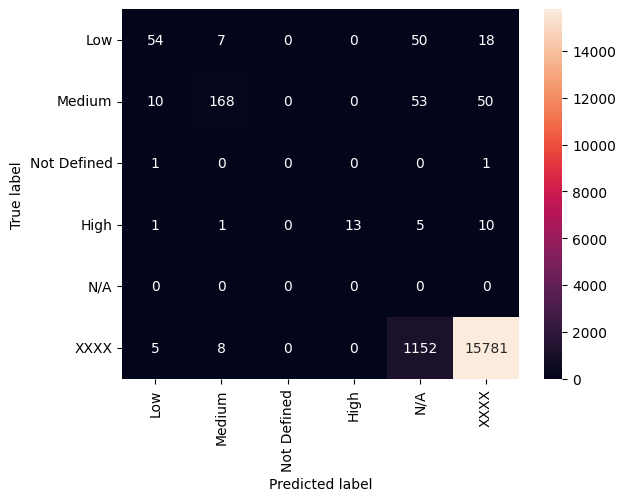}
        \caption{Availability Requirement}
    \end{subfigure}
    \hfill
    \begin{subfigure}[b]{0.3\textwidth}
        \includegraphics[width=\textwidth]{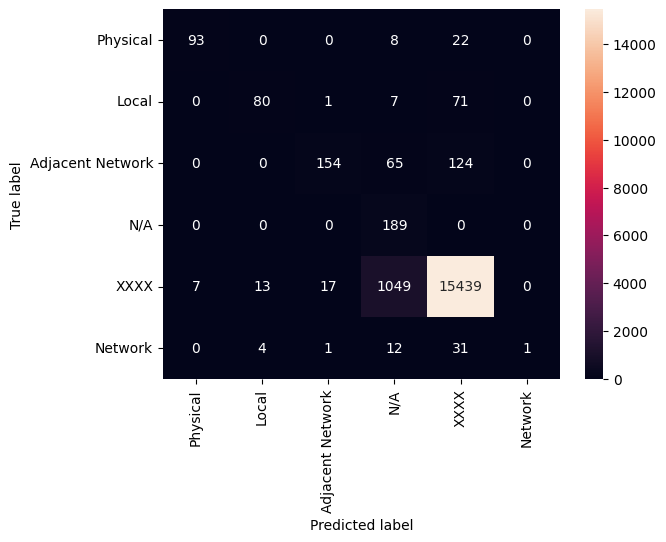}
        \caption{Modified Attack Vector}
    \end{subfigure}
    ~ 
    \begin{subfigure}[b]{0.3\textwidth}
        \includegraphics[width=\textwidth]{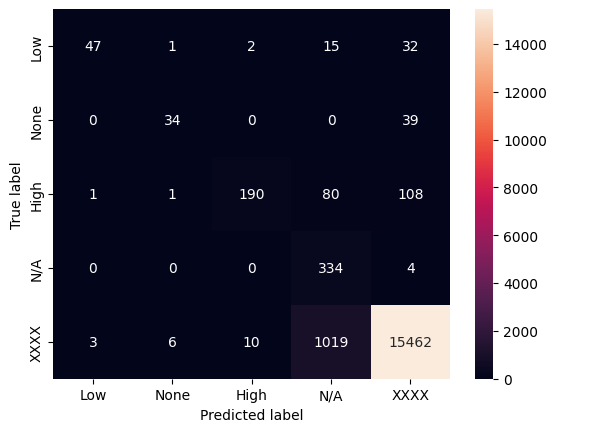}
        \caption{Modified Integrity}
    \end{subfigure}
    \caption{Environmental Vector confusion matrix: breakdown by metrics}\label{fig:env_vector}
\end{figure*}

\begin{table}[H]
      \begin{tabular}{ccl}
        \toprule
        \textbf{Environmental Metrics}&\textbf{Micro-F1}\\
        \midrule
        Modified Attack Complexity & 0.919\\
        Modified Availability & 0.923 \\
        Integrity Requirement & 0.922 \\
        Modified Attack Vector & 0.918\\
        Modified Confidentiality& 0.925\\
        Availability Requirement & 0.921\\
        Modified Scope & 0.924\\
        Modified User Interaction&0.922\\
        Modified Privileges Required&0.919\\
        Confidentiality Requirement&0.922\\
        Modified Integrity&0.924\\
      \bottomrule
    \end{tabular}
    \caption{Micro-F1 for Environmental metrics on test dataset (N=44K)}
      \label{tab:envmetric}
    \end{table}
The environmental metrics found in the test dataset and their corresponding micro-F1 are shown in \ref{tab:envmetric}. Figure \ref{fig:env_vector} shows the confusion matrix for these environmental metrics. Due to differences in CVSS versions, there are some environmental metrics that are not applicable to certain evaluations, and their values are represented as `N/A'. The value `XXXX' for environmental metrics refers to the cases where the expert did not use any value for that environmental metrics.

\section{Common errors in generation with CVE-LLM}
The common errors found in model outputs are mostly related to 
\begin{inparaenum}
    \item Wrong generation,
    \item Omission of key information, 
    \item Wrong key information, and
    \item Addition of spurious details 
\end{inparaenum}
Examples of these errors are shown in Table \ref{tab:generr} and \ref{tab:addition}.

We noticed that most of the errors in generation are due to the probabilistic regurgitation of training text, and there is less semantic understanding of the paradigm. We surmise that this can be corrected post-hoc with verification of the entities involved in the generation, chain of thought prompts and reasoning modules.

\begin{table*}
    \begin{center}
    \begin{small}
    \begin{tabular}{m{0.98\linewidth}}
    \hline
    \multicolumn{1}{c}{Prompt}\\
    \hline
    
    Below is an instruction that describes a task, paired with an input that provides further context.
    Write a response that appropriately completes the request.

    \#\#\# Instruction: Generate internal comments.
    
    \#\#\# Input:
    
    Organization: \censor{DI-DnA}
    
    Software: \censor{Syngo Carbon Monitoring}
    
    Product: \censor{Syngo Carbon Monitoring VB12A}
    
    Notification:
    
    A vulnerability in the mod\_proxy\_cluster in the Apache server. This may allow a malicious user to add a script in the `alias' parameter in the URL to trigger the stored cross-site scripting  vulnerability. By adding a script on the alias parameter on the URL, it adds a new virtual host and adds the script to the cluster-manager page.
    
    Components present in software: HTTP Server (httpd) - Apache Software Foundation - 2.4.57
    
    Base and Temporal Vectors: Attack Vector is Network. Attack Complexity is Low. Privileges Required is Low. User Interaction is Required. Scope is Changed. Confidentiality is Low. Integrity is Low. Availability is None. Exploit Code Maturity is Unproven. Remediation Level is Unavailable. Report Confidence is Reasonable.

    CVSS Version: 3.1

    \#\#\# Response:
    \\
    \hline
    \multicolumn{1}{c}{Expected Response}\\
    \hline
    Component will be updated in future release of product.\\
    \hline
    \multicolumn{1}{c}{Generated Response}\\
    \hline
    Component is updated to the latest version of GEC agent.\\
    \hline
    \hline
    \end{tabular}
    \end{small}
    \end{center}
    \caption{
    Addition of unrelated incorrect information}
    \label{tab:addition}
    \end{table*}

\begin{figure*}
    \centering
    \begin{subfigure}[b]{0.24\textwidth}
        \includegraphics[width=\textwidth]{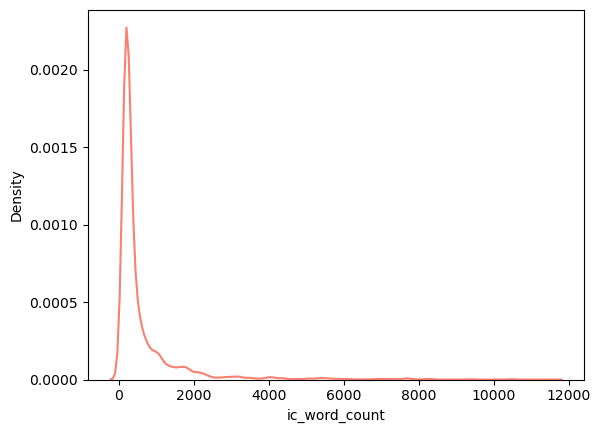}
        \caption{Training instruction Word count density plot}
    \end{subfigure}
    ~ 
    \begin{subfigure}[b]{0.24\textwidth}
        \includegraphics[width=\textwidth]{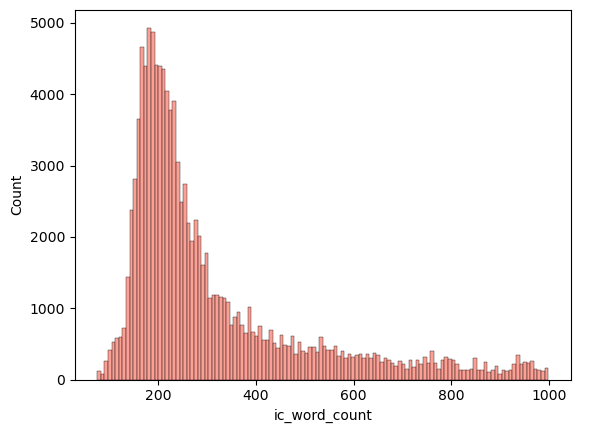}
        \caption{Train instruction Word count histogram for instructions with <1k words}
    \end{subfigure}
    ~ 
    \begin{subfigure}[b]{0.24\textwidth}
        \includegraphics[width=\textwidth]{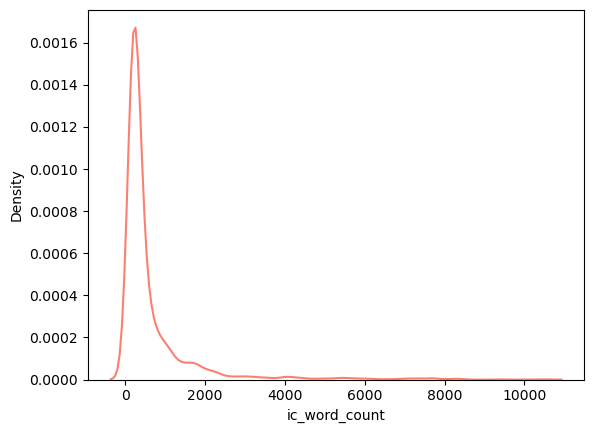}
        \caption{Test instruction Word count density plot}
    \end{subfigure}
     \begin{subfigure}[b]{0.24\textwidth}
        \includegraphics[width=\textwidth]{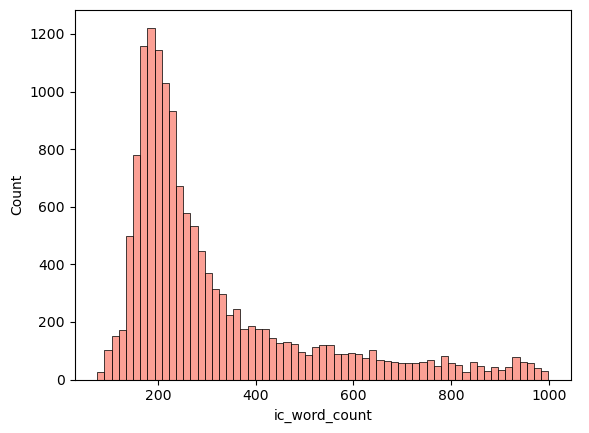}
        \caption{Test instruction Word count histogram for instructions with <1k words}
    \end{subfigure}
    \hfill
    ~ 
    \begin{subfigure}[b]{0.24\textwidth}
        \includegraphics[width=\textwidth]{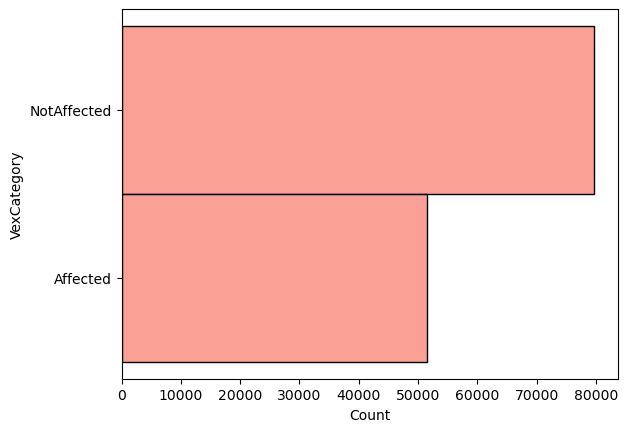}
        \caption{Training dataset \VexCategory histogram}
    \end{subfigure}
    \begin{subfigure}[b]{0.24\textwidth}
        \includegraphics[width=\textwidth]{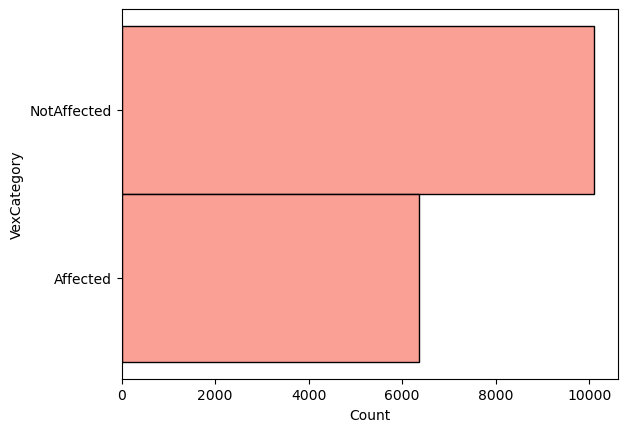}
        \caption{Test dataset \VexCategory histogram}
    \end{subfigure}
    \begin{subfigure}[b]{0.24\textwidth}
        \includegraphics[width=\textwidth]{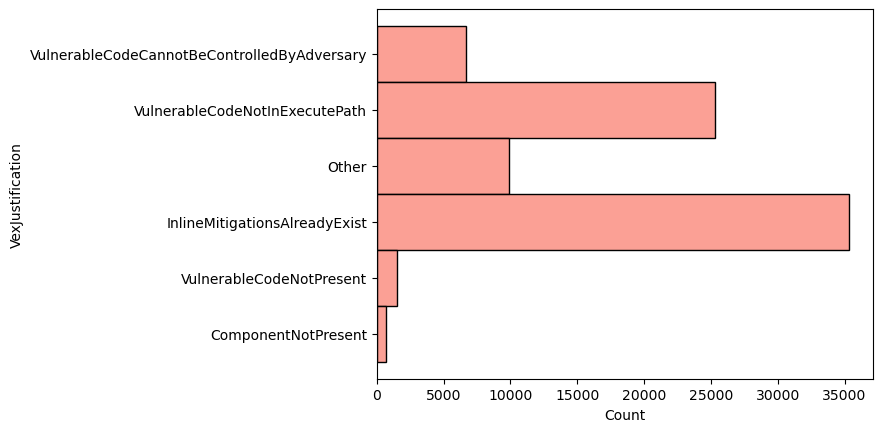}
        \caption{Training dataset \VexJustification histogram}
    \end{subfigure}
    \begin{subfigure}[b]{0.24\textwidth}
        \includegraphics[width=\textwidth]{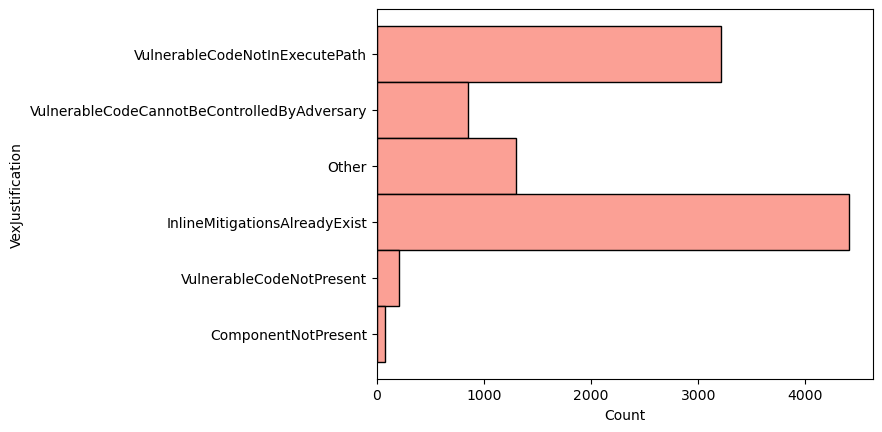}
        \caption{Test dataset \VexJustification histogram}
    \end{subfigure}
    \caption{Dataset statistics}\label{fig:data_stats}
\end{figure*}

\end{document}